\documentclass[sn-basic,Numbered]{sn-jnl}% Basic Springer Nature Reference Style/Chemistry Reference Style
\usepackage{graphicx}%
\usepackage{multirow}%
\usepackage{amsmath,amssymb,amsfonts}%
\usepackage{amsthm}%
\usepackage{mathrsfs}%
\usepackage[title]{appendix}%
\usepackage{xcolor}%
\usepackage{textcomp}%
\usepackage{manyfoot}%
\usepackage{booktabs}%
\usepackage{amssymb}
\usepackage{xcolor}
\usepackage{hyperref}
\usepackage{makecell}
\usepackage{url}
\usepackage{rotating}
\usepackage{adjustbox} 
\usepackage{multirow,multicol}
\usepackage{subfigure}
\usepackage{booktabs}
\usepackage{makecell}
\usepackage{xcolor}
\usepackage{tabularx}

\newcommand{\rot}[1]{\rotatebox[origin=c]{90}{#1}}
\let\cline\cmidrule
\usepackage[switch]{lineno}
\usepackage{multirow,multicol}

\theoremstyle{thmstyleone}%
\theoremstyle{thmstyletwo}%

\theoremstyle{thmstylethree}%

\begin{document}

\title[Matching Tasks to Objectives]{Matching Tasks to Objectives: Fine-Tuning and Prompt-Tuning Strategies for Encoder-Decoder Pre-trained Language Models}

%\title[Matching Tasks to Objective: Task-Oriented Strategies for Fine-Tuning and Prompt-Tuning of Encoder-Decoder Pre-trained Language Models]{Matching Tasks to Objective: Task-Oriented Strategies for Fine-Tuning and Prompt-Tuning of Encoder-Decoder Pre-trained Language Models}

%\title[From Pretraining to Fine-Tuning: Task-Oriented Strategies for Fine-Tuning and Prompt-Tuning]{From Pretraining to Fine-Tuning: Task-Oriented Strategies for Fine-Tuning and Prompt-Tuning}

%\title[Task-Oriented Adaptation of PLMs: Insights from Fine-Tuning and Prompt-Tuning]{Task-Oriented Adaptation of PLMs: Insights from Fine-Tuning and Prompt-Tuning}

%\title[Matching Pre-trained Language Models to Specific Tasks: Fine-tuning and Prompt-tuning Strategies]{Matching Pre-trained Language Models to Specific Tasks: Fine-tuning and Prompt-tuning Strategies}

%%=============================================================%%
%% Prefix	-> \pfx{Dr}
%% GivenName	-> \fnm{Joergen W.}
%% Particle	-> \spfx{van der} -> surname prefix
%% FamilyName	-> \sur{Ploeg}
%% Suffix	-> \sfx{IV}
%% NatureName	-> \tanm{Poet Laureate} -> Title after name
%% Degrees	-> \dgr{MSc, PhD}
%% \author*[1,2]{\pfx{Dr} \fnm{Joergen W.} \spfx{van der} \sur{Ploeg} \sfx{IV} \tanm{Poet Laureate} 
%%                 \dgr{MSc, PhD}}\email{iauthor@gmail.com}
%%=============================================================%%

\author*[1]{\fnm{Ahmad} \sur{Pouramini}}\email{ahmad.pouramini@ut.ac.ir}
%\orcidlink{0000-0002-3652-0935}

\author[1]{\fnm{Hesham} \sur{Faili}}\email{hfaili@ut.ac.ir}
\equalcont{These authors contributed equally to this work.}

\affil*[1]{\orgdiv{School
		of Electrical and Computer Engineering, College of
		Engineering}, \orgname{University of Tehran}, \orgaddress{\street{North Kargar Street}, \city{Tehran}, \postcode{515-14395}, \state{Tehran}, \country{Iran}}}

%%==================================%%
%% sample for unstructured abstract %%
%%==================================%%

\abstract{
Prompt-based learning has emerged as a dominant paradigm in natural language processing. This study explores the impact of diverse pre-training objectives on the performance of encoder-decoder pre-trained language models across generation and question answering tasks, with a focus on commonsense knowledge retrieval and completion. We highlight the benefits of incorporating multiple objectives during both pre-training and fine-tuning stages. We introduce the Match Task to Objective (MTO) framework and methods for determining the appropriate objective for a given task. This framework offers automated methods to prepare task-related data for adaptation through unsupervised training, based on the identified objective. In the fine-tuning stage, we design novel templates that align with the objectives of the pre-training and adaptation stages. When aligned with task requirements, these strategies can achieve a performance gain of over 120\% compared to conventional methods in few-shot settings. They significantly outperform related works in few-shot settings and exceed the baseline even in full-dataset scenarios.
Furthermore, we extend this approach to include prompt-tuning methodologies, providing guidance for more effective soft prompt engineering and optimization. Our strategies significantly enhance prompt-tuning performance as well. These insights hold substantial value, precisely guiding the selection and optimization of models customized for specific tasks. Code is available at  \url{https://github.com/puraminy/MTO/}
}


\maketitle

\section{Introduction}
In the rapidly evolving field of natural language processing (NLP), prompting methods have garnered considerable attention due to their potential to effectively leverage the capabilities of language models \citep{ppp}. In this novel paradigm, the traditional method of customizing pre-trained language models (PLMs) for specific downstream tasks using focused objective engineering gives way to a process where the downstream tasks are restructured to resemble those encountered during the original LM training. This transformation becomes achievable by strategically integrating textual prompts \citep{ppp}. 

For instance, consider the task of recognizing the sentiment of a social media post, such as "I missed the train today". Traditionally, this might involve explicit task-specific adaptations of the LM. However, within this new framework, extending the text with a prompt like "I felt so ---", prompts the LM to predict and fill in the blank with an emotion-laden word. This approach exemplifies how prompting can guide the LM to infer emotions even from relatively unremarkable statements. 

Knowledge-base (KB) completion is another application of PLMs where prompts could be employed to infer missing information within a knowledge graph. However, it's essential to note that different relations within the knowledge graph may require distinct techniques and approaches to accurately predict completions. For instance, factual or one-to-one relations may differ significantly from the many-to-many relations found in commonsense knowledge graphs or those between events and concepts \cite{lama, aser, springer-event, concept}. These variations necessitate tailored strategies for effectively leveraging PLMs in KB completion tasks.

The advantage of this prompting is that with a collection of appropriate prompts, a single PLM can be utilized to address multiple tasks \citep{brown}. However, it's imperative that the model has undergone appropriate pre-training to proficiently cater to the diverse requirements of various tasks.

In this study, our focus revolves around identifying factors that can augment the performance of a PLM on various tasks within few-shot settings. Our underlying conjecture is that a fine-tuning methodology closely aligned with the unsupervised approach, has the potential to enhance performance under specific circumstances.

To achieve this aim, we evaluate the efficacy of several pre-training objectives and associated prompting templates through fine-tuning. Each configuration is designed to align with the unsupervised objectives established during the model's pretraining phase. Additionally, we introduce methods to prepare task-related data for complementary unsupervised training using the objective suitable for a task. These objectives can be determined by considering the nature of the task or generated by a classifier we designed for this purpose.

\subsection{Research Objectives}
In summary, our research addresses the following key inquiries:
\begin{itemize}
	
	\item We aim to understand the influence of pre-training objectives on the performance of a PLM within specific tasks, with a primary focus on enhancing commonsense knowledge generation through the use of  encoder-decoder models.
	
	\item We explore the effects of aligning the input format of a language model with its unsupervised pre-training objectives. Specifically, we investigate which tasks benefit most from this alignment and aim to establish a framework for linking tasks to specific pre-training objectives and PLMs. Additionally, we provide methods to generate task-related data for extended unsupervised training to further enhance performance.
	
	\item We investigate whether alignment strategies originally designed for KB completion tasks can be effectively applied to question answering tasks. Despite the distinct nature of these tasks, we propose novel alignment methods and demonstrate that these strategies can still produce favorable outcomes and boost performance.
	
	\item Building on our earlier investigations, we further explore the applicability of our findings to prompt tuning methodologies. We seek to determine if the insights garnered can be extended to enhance task performance through prompt engineering.
	
\end{itemize}

\section{Related Works}
Prompts have found valuable applications in knowledge retrieval from pre-trained language models. Notably, Petroni et al. conducted a study where they explored the direct retrieval of factual and commonsense knowledge from PLMs without fine-tuning \citep{lama}. This was achieved by converting queries into cloze-style prompts and asking the model to complete them. Their research revealed that this technique is particularly effective for retrieving one-to-one factual knowledge. However, when it comes to completing commonsense declarative relationships, PLMs encounter limitations in effectively extending their language modeling capabilities to this task \citep{lama,rel-yin,rel-zhou,davison-cs-mining,rel-cs}. 

Following this, a multitude of other studies in the realm of knowledge probing have endeavored to devise more efficient prompts or probing techniques for extracting factual knowledge \citep{rel-shin, rel-zhao, rel-jiang, k-survey}.

The works that are more related to our paper are those aimed at improving knowledge extraction from PLMs through either prompt tuning or fine-tuning. The research on prompt tuning includes the extraction of prompts from the web \citep{rel-jiang}, the optimization of prompts within the discrete domain of words and tokens \citep{rel-shin}, or the optimization of prompts within the continuous embedding space \citep{rel-contprompt,rel-contprompt2}. %Nonetheless, these endeavors predominantly target factual knowledge.

A number of works have addressed the task of fine-tuning a PLM by utilizing tuples from established datasets or knowledge graphs \citep{comet, comet2020, analyzingCS,rel-fin1,rel-fin2, west}.  \citep{rel-fin1} demonstrates that PLMs fine-tuned on a specific set of queries outperform other knowledge probing techniques. 
However, other research questions the efficiency of fine-tuning, asserting that it could potentially result in the loss of pre-trained knowledge and capabilities \citep{rel-wallet, rel-fin2,rel-cao}.  

%Furthermore, they highlighted that performance hinges on the alignment between the distribution of training data and test data, which could potentially result in an overestimation of test data performance \citep{rel-cao, rel-fin2}. 

\citep{rel-fin2} demonstrate that while fine-tuning enhances task learning, it can degrade performance if there's a significant mismatch between training and test datasets. They propose using an ensemble of fine-tuned and zero-shot models or incorporating pre-training objectives during fine-tuning to mitigate this issue. \citep{rel-wallet} highlight the importance of pre-training objectives, finding that ranking objectives better preserve factual knowledge compared to masked language modeling and question-answering, though masked models are better at acquiring new knowledge from the training data. However, their work focused on encoder-only language models like BERT.

\cite{analyzingCS} studied knowledge emergence in commonsense knowledge models, which are generative and typically encoder-decoder models. They found that these models can quickly adapt from limited examples, suggesting that knowledge graph fine-tuning effectively learns an interface to the knowledge encoded during pretraining. They observed more parameter shift occurs in the decoder rather than the encoder during fine-tuning, implying that much of the parameter shift may be due to learning how to express commonsense knowledge in a declarative form.

Some studies have attempted to enhance the zero-shot performance of PLMs through transfer learning by fine-tuning a model on a selection of relevant tasks in a multi-task fashion before adapting it to a target task with limited labeled samples \cite{t0, rainbow, unified}. The authors of \cite{t0} propose a fine-tuned encoder-decoder language model based on Google's T5 \cite{t5} architecture, trained on a diverse set of 50 datasets using a prompted format to improve zero-shot performance on unseen tasks.

\cite{rainbow} focus on transfer learning via multitask training with an emphasis on general common-sense knowledge, evaluating its application to common-sense QA tasks. They concluded that the best results are achieved by sequentially training on multiple datasets (excluding the target dataset) through multitask training, followed by continued training on the target dataset. They also found that transferring knowledge from graphs to QA tasks has little impact due to the differences in task formats and the distinction between generative and discriminative approaches. \cite{unified} attempted to build a model based on T5, fine-tuned using diverse QA tasks by unifying the format of these tasks.

There has also been research on continued pre-training in PLMs, with focus on model adaptation: either adapting data \cite{dontstop, springer-cs}, or adapting the training objective towards downstream tasks \cite{related-what}.

\cite{dontstop} show that continuing pretraining on a task corpus augmented with simple data selection strategies is effective in enhancing task performance, especially when resources for domain-adaptive pretraining are unavailable.

\cite{related-what} conducted experiments on various combinations of model architectures and training objectives. They found that while a decoder-only model achieves the best zero-shot performance immediately after pretraining, an encoder-decoder model with masked language modeling performs best once multitask fine-tuning is applied. However, since multitask fine-tuning may not be suitable for many open-ended generative tasks, they proposed extending unsupervised pretraining with different architectures and/or objectives as a practice for adaptation.

In this paper, we address the adaptation of both data and objectives toward specific tasks, offering methods to align the training data format and objectives with pre-trained models, as well as methods to provide data for extended unsupervised training (adaptation phase). We focus on generative encoder-decoder models, known for their ability to generate extended responses and handle various tasks \citep{t0, rel-timo}. However, these ideas can be extended to other types of models and tasks. We consider the requirements of each task and conduct experiments to determine the most suitable objectives. Our emphasis shifts to commonsense knowledge, where both full-data and few-shot fine-tuning have demonstrated significant benefits \citep{comet2020, analyzingCS}.  

\section{Background}
To develop an understanding of how PLMs undergo their training process, we begin by reviewing the prevalent pretraining objectives employed in these models. 
\subsection{Pre-training Objectives in Language Models}
Pre-trained language models initially undergo unsupervised pre-training, where they are exposed to extensive unlabeled text data. The fundamental training objective for these models often revolves around predicting the probability of text segments.

Diverse techniques can be employed to structure sentences as inputs and outputs for the model. In the prevalent \textbf{Language Model objectives}, text prediction usually occurs in an autoregressive manner, sequentially predicting tokens in the sequence. 
As an alternative, \textbf{denoising objectives} introduce noise to the input sentence, such as permutation, deletion, or word masking, and subsequently aim to predict the original unnoised sentence.

Accordingly, language models (LMs) can be broadly categorized based on their objectives. \textbf{Left-to-Right LMs} are a variant of auto-regressive LM that generate text by predicting tokens in a left-to-right sequence, proving optimal for tasks demanding coherent text generation. For tasks requiring a profound understanding of context, such as classification and question answering, \textbf{Masked LMs} are more effective. The objective of these models is to predict concealed text fragments, utilizing contextual cues from surrounding words \cite{ppp}.

In the realm of conditional text generation, two prominent architectures emerge: \textbf{Prefix LMs} generate target text by conditioning the prediction process on a provided prefix sequence, often mirroring the input sequence. The prefix is encoded using the same model parameters, but with the addition of a fully-connected mask mechanism. \textbf{Encoder-Decoder LMs}, akin to Prefix LMs, generate target text using a left-to-right approach. However, they distinguish themselves by incorporating a separate encoder to process the input sequence effectively. This encoder handles input sequences of varying complexities, rendering these models indispensable in a wide array of natural language processing tasks.

\begin{table}
	\centering
	\begin{tabular}{|p{3cm} | c | c| }
		\hline
		\textbf{Objective} & {\textbf{Input}}  & {\textbf{Target}} \\ 
		\hline
		Prefix Language Modeling &  \makecell[l]{Alice cooks \\ Before cooking one needs }  &  \makecell{to satisfy  hunger  \\ to prepare ingredients}  \\
		\hline
		\multirow{2}{3cm}{I.i.d Noise Replace Spans}   &  \makecell[l]{ Alice cooks  \textcolor{brown}{\textbf{X}}.  Before \\ that she needs \textcolor{brown}{\textbf{Y}}} & \makecell[l]{\textcolor{brown}{\textbf{X}} to satisfy hunger \\ \textcolor{brown}{\textbf{Y}} to prepare ingredients}   \\ \cline{2-3} 
		& Alice  cooks \textcolor{brown}{\textbf{X}} meals &  \makecell[l]{\textcolor{brown}{\textbf{X}} delicious}  \\
		\hline		
	\end{tabular}
	\caption{Examples of inputs and outputs for two unsupervised  objectives in an encoder-decoder model.}
	
	\label{table:objectives}
	
\end{table}

Given that our focus lies in conditional text generation tasks, this paper primarily concentrates on objectives frequently utilized in Encoder-Decoder models.
\subsubsection{Objectives for Encoder-Decoder Models}
Table \ref{table:objectives} presents examples of two specific objectives utilized in encoder-decoder models, which are further explored in this study.
\begin{itemize}
	\item \textbf{Prefix Language Modeling}
	In the fundamental "prefix language modeling," a segment of text is randomly divided into prefix and target segments. The prefix is used as input to the encoder, while the target segment becomes the output sequence.
	\item \textbf{Denoising Objective}:
	In the denoising objective, segments of the input sequence are randomly selected and concealed using  unique mask tokens. The target is then composed of the concealed spans of tokens, bordered by the same mask tokens employed in the input sequence. Through this process, the model learns to restore the initial sequence by forecasting the masked-out spans of tokens.	
\end{itemize}

\subsection{Prompt-Tuning}
Prompt-tuning is a parameter-efficient strategy to enhance a PLM by refining its response to specific textual prompts. Unlike traditional fine-tuning, which adapts the entire model for a new task, prompt-tuning exclusively adjusts the prompts themselves \citep{ppp}.
Prompt optimization can take place in either the discrete space of words or the continuous embedding space of the model. 
In this paper, we utilized gradient descent over the continuous space of model embeddings to optimize the prompts \citep{lester2021power}. The input embeddings, combined with the prompt embeddings as $[\mathbf{P}; \mathbf{X}]$, are subsequently input into the frozen LM to maximize the likelihood of generating the desired target sequence $\mathbf{y}$. Formally, given the pre-trained language model (LM) parameters $\theta$, this optimization is expressed as:

\[
\max_{\mathbf{P}} p_{\theta}(\mathbf{y} | [\mathbf{P};\mathbf{X}]),
\]

where $\mathbf{P} = [h_1, \ldots, h_m] \in \mathbb{R}^{m \times d}$, with $m$ representing the prompt's length and $d$ representing the LM embedding dimension.

\begin{figure}[t]
	\centering
	\includegraphics[width=0.9\linewidth]{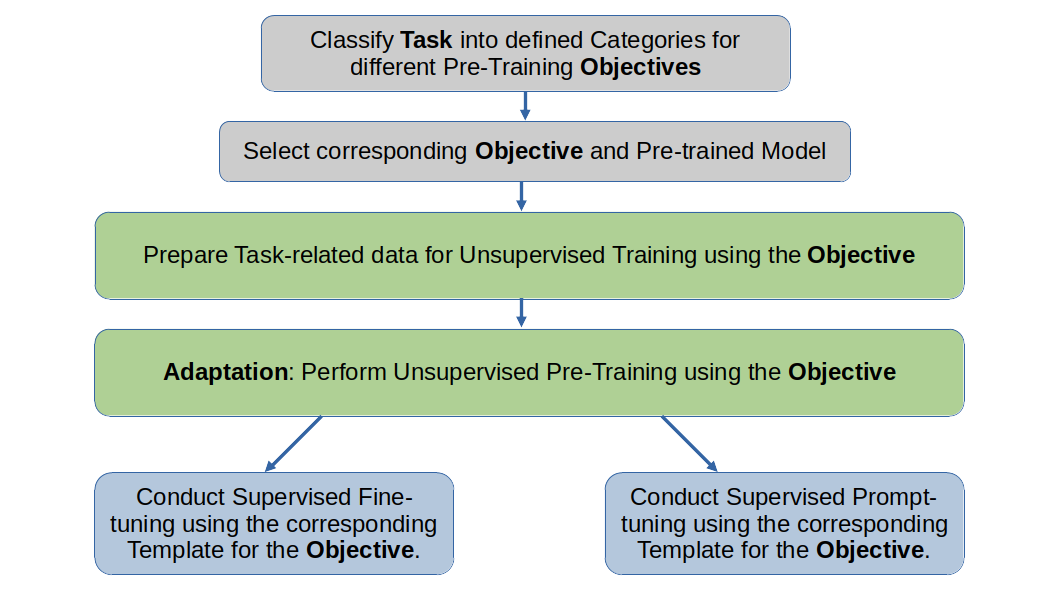}
	\caption[Overal framework of our proposed framework MTO]{General steps of our proposed framework MTO}
	\label{fig:overal}
\end{figure}

\section{Method}
Figure \ref{fig:overal} illustrates the general steps of our proposed framework MTO. Initially, we classify a task to certain task categories to identify the appropriate objective for training the model. This classification can be performed either by an expert familiar with the task or by using a classifier.

Once the task is classified, we select a pre-trained language model (PLM) that aligns with the identified objective based on the main objectives used in the model's pre-training phase. To further adapt the selected language model to the specific task, we continue unsupervised pretraining using the identified objective and task-aligned data. The data can be prepared based on the target objective using automatic methods, which we describe in the following sections.

In the subsequent stage, we fine-tune the model on task-specific samples, employing objectives and templates that match the initial objective. Alternatively, prompt-tuning can be utilized as a more parameter-efficient approach. The following sections provide a detailed description of each stage, including our task classifications and corresponding templates, with a particular emphasis on knowledge retrieval from pre-trained models.

\subsection{Task Categories}
Table \ref{table:dataset} presents the task categories employed in our experimental analysis, showcasing examples of the input and output for selected tasks in each category. We conduct experiments on three distinct categories of tasks:

\begin{itemize}
	\item \textbf{Mask-Filling:} These tasks primarily pertain to knowledge retrieval or knowledge base completion. In such cases, a well-constructed prompt in conjunction with a masked token can potentially assist in identifying a plausible tail, given a head and relation. The targets are often one or few words. 
	
	\item \textbf{Map-Phrasal:} These tasks often involve mapping or learning relationships between a head and tail, and then generalizing that understanding to novel instances. In these scenarios, prompts might aid the process, but additional examples are likely necessary to fully grasp the relationship. Moreover, the target is often a full sentence or a sentence fragment.
		
	\item \textbf{Question-Answering (QA):} These tasks involve providing a specific answer to a given question based on a provided context or knowledge base. They could be free or multi-choice questions. The targets in QA tasks are often a span of text, a single word, or a short phrase. Prompting can enhance performance in QA tasks by guiding the model to focus on relevant parts of the context and by structuring the questions and answers in a way that is easier for the model to understand and process.

	%\item \textbf{Classification:} Classification tasks primarily revolve around discerning the class of an input that may comprise one or several components. While prompting can be beneficial for these tasks, we expect a need for a deeper understanding of input variability across multiple examples. 
 
\end{itemize}

	\begin{table*}
	\renewcommand{\arraystretch}{1.3}
	\centering
	\begin{tabular}{c|c|c|c}
		\hline
		\textbf{Group} & \textbf{Input} & \textbf{Task} & \textbf{Target} \\
		\hline
		\multirow{7}{*}{\makecell{Mask \\ Filling}} 
		& \multirow{3}{*}{bread} & ObjectUse & make french toast \\
		& & AtLocation & basket; pantry \\
		& & HasProperty & cooked; nice to eat \\
		\cline{2-4}
		& \multirow{1}{*}{baker} & CapableOf & coat cake with icing \\
		\cline{2-4}
		& \makecell{PersonX cooks \\ --- meals} & isFilledBy & delicious; nutritious \\
		\cline{2-4}
		& PersonX teaches & xAttr & knowledgeable; educated \\
		\midrule
		\multirow{3}{*}{\makecell{Map \\ Phrasal}} 
		& \multirow{3}{*}{PersonX cooks} & xIntent & to satisfy hunger \\
		& & xNeed & to prepare ingredients \\
		& & xWant & to eat \\
		\midrule
		\makecell{Question \\ Answering} 
		& \makecell{Where do you put \\ food to keep it cold?} & \makecell{1. Oven \\ 2. Refrigerator} & Choice2 \\
		\bottomrule
	\end{tabular}
	\vspace{10pt} % Add extra space before the caption
	\caption{Selected relations from $ATOMIC^{20}_{20}$, along with illustrative examples, are categorized into Mask-Filling and Map-Phrasal groups based on their structure and difficulty}
	\label{table:dataset}
\end{table*}

\begin{figure}[h!]
	\centering
	\includegraphics[width=1\linewidth]{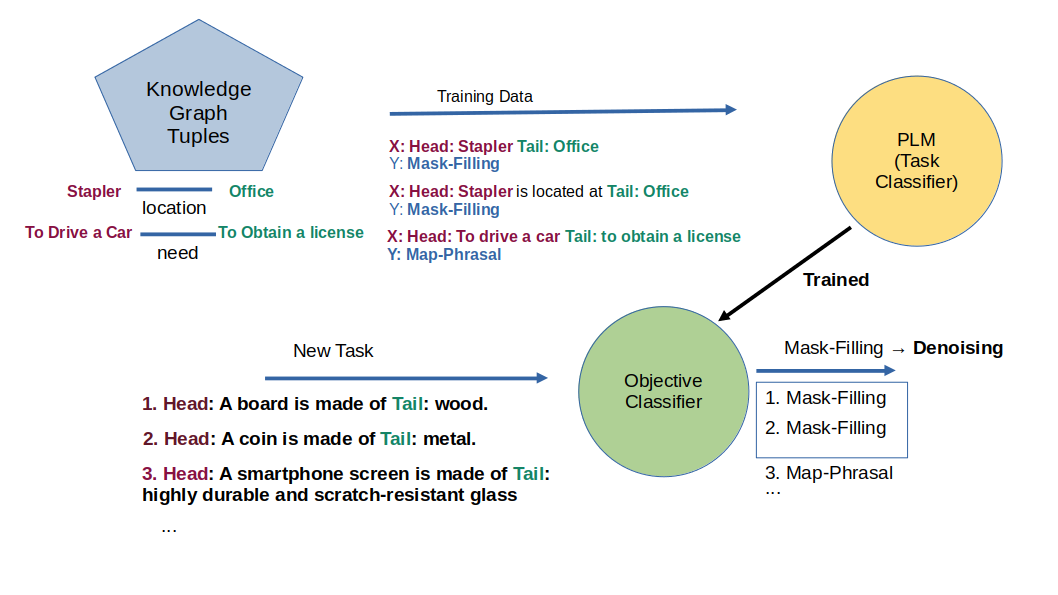}
	\caption{Classifier designed to associate a given task with a category. The category is then mapped to the training objective.}
	\label{fig:classifier}
\end{figure}

\subsection{Classify Task into Defined Categories} \label{sec:classifier}
To identify the appropriate training objectives and stages, we can categorize given tasks. Along with QA category, we can define categories for generative tasks. As shown in Table \ref{table:dataset}, examples of various tasks can be categorized based on their nature or the similarity of methods used to solve them. The category of a task can be determined by an expert or derived from examples of the tasks themselves. For instance, the tail of Mask-filling tasks often involves a short word that can fill a masked word within a prompt template designed for the task (e.g., "\textbf{Stapler} is located at the \textbf{office}"). These sentences are more likely to be found in natural texts and are well-suited for the denoising objective using masked tokens. For Map-Phrasal tasks, we may design a template with two parts that can be broken down (e.g., "\textbf{To cook}, one needs \textbf{to first prepare ingredients}"), which typically requires more samples to be effectively understood by the model and align better with the language model objective. 

As an automatic solution, once we identify and categorize several tasks, we can train a classifier using samples of these tasks to categorize new tasks. Figure \ref{fig:classifier} demonstrates such a classifier. The classifier is trained on the head-tail pairs of sample tasks along with their assigned categories as labels. If we have many relations, we can convert the relations to natural phrases and use them in continuation of the head in the input. After training the classifier, we apply it to samples of a new task. The model classifies each sample into one of the categories, and the category assigned to the majority of samples is selected as the category for the task. This category is then mapped to the training objective.

\subsection{Select Corresponding Objective and Pre-trained Model}
Based on the category of a given task, we associate the task with an objective that suits its type. As mentioned above, we select the denoising objective for the Masked-Filling category and the language model (LM) objective for the Map-Phrasal category. For example, among the different versions of T5, we could select T5-LM, T5-v1, or T5-base, which employ either or both of these objectives in their pre-training stage.

\begin{figure}[bh!]
	\centering
	\includegraphics[width=1\linewidth]{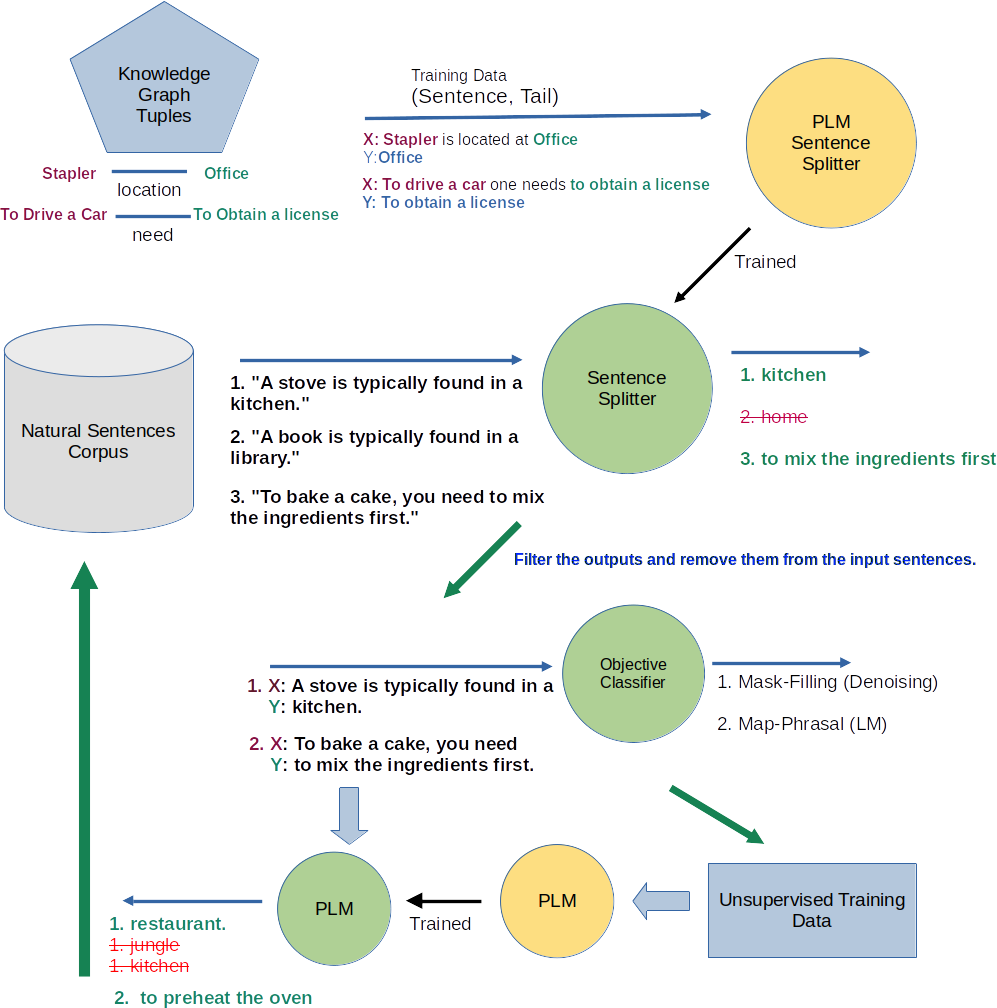}
\caption{Overview of the process to prepare data for unsupervised pre-training before fine-tuning the model on a specific task.}
	\label{fig:unsup}
\end{figure}

\subsection{Adaptation: Unsupervised Pre-Training Using the Objective and Task-related data}\label{sec:splitter}
Before fine-tuning the selected model on a specific task, we can first train it using the target objective or a blend of objectives on natural sentences in an unsupervised manner. Ideally, we use a corpus of natural sentences related to our desired tasks. Each sentence is divided into two parts: the model is trained to generate the second part based on the first, or to fill in omitted parts of the sentence. We designed a process to provide training data for this phase, as illustrated in Figure \ref{fig:unsup}. To determine sentence splits, we train a model known as the Sentence Splitter using tuples extracted from a knowledge graph. This model learns to segment natural sentences into components analogous to the head and tail of a knowledge graph.

Our hypothesis is that breaking sentences this way is more efficient for distilling knowledge from sentences and can also enhance the performance of supervised training by adopting a similar strategy for formatting the model's input and output.

To achieve this, we use templates to convert the knowledge graph tuples into natural sentences. These sentences serve as the input for the model, with the tail specified as the target. By training the model this way, it learns to divide new sentences into similar parts. This model is then applied to natural sentences to split them and provide the necessary data for our unsupervised training. We can then train the model using a denoising objective by replacing the second part with a mask token, or using a language modeling (LM) objective without mask tokens.

Optionally, we can use a classifier to determine the suitable objective for each produced sample (as described in Section \ref{sec:classifier}). This adds another label to our split sentences, specifying whether the sample is suitable for the denoising objective or the LM objective. We call data provided in this manner "Mixed" data.

After training the model on the data provided through this stage, it can then be fine-tuned on task-specific samples using the templates designed for each objective. 

As depicted in Figure \ref{fig:unsup}, it is important to note that applying the trained model to its input data at this stage may generate new completions for the first part of each sentence. When combined, these completions can form new natural sentences, thereby enhancing our unsupervised data through a bootstrapping process.

The model is now prepared for the fine-tuning or prompt-tuning stage. In the subsequent phase, we create templates tailored to our tasks, ensuring alignment between the format and objectives of both pre-training and adaptation. We discuss this further in the next section. I

\subsection{Designing Fine-Tuning Template}
Based on the selected model, we design the corresponding template to format the input and output according to the chosen objective. Below, we present the templates we devised to format the input and output of a PLM for both fine-tuning and prompt-tuning. These setups are customized in alignment with the objectives outlined earlier, particularly suited for encoder-decoder models.

%\documentclass{article}
%\usepackage[table]{xcolor}
%\usepackage{multirow}
%\usepackage{xcolor}
%\usepackage{booktabs}

%\begin{document}
	
	\begin{table*}[th!]
		\centering
		\begin{tabular}{l | l | l}
			\toprule
			\multicolumn{3}{c}{\textbf{Fine-Tuning}} \\
			\midrule
			\multirow{1}{*}{\textbf{Template}} &  \\
			& \textit{Input} & \textit{Target} \\
			\hline
			Mapping & She cooks & to satisfy hunger \\
			\hline
			Prompting & She cooks \textcolor{cyan}{because she intends} & to satisfy hunger \\
			\hline
			MaskedMapping & She cooks \textcolor{brown}{\textbf{X}} & \textcolor{brown}{\textbf{X}} to satisfy hunger \\
			\hline
			MaskedPrompting & She cooks \textcolor{cyan}{because she intends} \textcolor{brown}{\textbf{X}} & \textcolor{brown}{\textbf{X}} to satisfy hunger \\
			\midrule
			\multicolumn{3}{c}{\textbf{Prompt-Tuning}} \\
			\midrule
			PrePT & \textcolor{red}{$ \mathbf{P}_1 \mathbf{ P}_2  \dots  \mathbf{P}_n $} She cooks \textcolor{cyan}{ because she intends} & to satisfy hunger \\
			\hline
			PostPT & She cooks \textcolor{red}{$ \mathbf{P}_1 \mathbf{ P}_2  \dots  \mathbf{P}_n $} & to satisfy hunger \\
			\hline
			MaskedPrePT & \textcolor{red}{ $ \mathbf{P}_1 \mathbf{ P}_2  \dots  \mathbf{P}_n $} She cooks \textcolor{cyan}{ because she intends} \textcolor{brown}{\textbf{X}} & \textcolor{brown}{\textbf{X}} to satisfy hunger \\
			\hline
			MaskedPostPT & She cooks \textcolor{red}{$ \mathbf{P}_1 \mathbf{ P}_2  \dots  \mathbf{P}_n $} \textcolor{brown}{\textbf{X}} & \textcolor{brown}{\textbf{X}} to satisfy hunger \\
			\hline
		\end{tabular}
		\caption{Templates for Formatting the Input and Output of an Encoder-Decoder Model during Fine-Tuning and Prompt-Tuning for Knowledge Completion Tasks.}
		\label{table:methods}
	\end{table*}
	
%\end{document}

\subsubsection{Templates for Fine-Tuning}
Table \ref{table:methods} outlines various methods employed for structuring the model's input and output during fine-tuning. These methods originate from the supervised or unsupervised objectives:

\begin{itemize}
	\item  \textbf{Mapping:}
	This approach corresponds to the traditional supervised method, involving fine-tuning through supervised sequence-to-sequence techniques. In cases of multi-tasking, a prefix is appended to each input to match the respective task.
	\item  \textbf{Prompting:}	
This technique is similar to the previously mentioned method but structures input tuples as natural language prompts. We expect it to align well with the language model's objectives and be particularly suitable for longer text complements, such as the xNeed relation in Table \ref{table:dataset}.

	\item  \textbf{MaskedMapping:}
	Drawing inspiration from the unsupervised denoising objective, this template utilizes a unique mask token (referred to as "X") as a placeholder for the target element.  This token is required to come before the target fragment in the output sequence. Typically, the number of fragments and their corresponding mask tokens can vary. This approach is particularly effective for predicting short missing fragments..
	
	\item \textbf{MaskedPrompting:}
	Combining aspects of both the Prompting and MaskedMapping, this template transforms inputs into natural language prompts and replaces the target with a placeholder or masked token. This fusion enables the generation of a coherent natural language sentence by combining the input and the replaced target. We expect that this template is well-suited for generating shorter complements, such as AtLocation in Table \ref{table:dataset}.
	
\end{itemize}

%\documentclass{article}
%\usepackage[table]{xcolor}
%\usepackage{multirow}
%\usepackage{xcolor}
%\usepackage{makecell}
%\usepackage{booktabs} % for \midrule

%\begin{document}
	
	\begin{table*}[th!]
		\centering
		\begin{tabular}{l | l | l}
			\toprule
			\multicolumn{3}{c}{\textbf{QA Fine-Tuning}} \\
			\midrule
			\multirow{1}{*}{\textbf{Template}} &  \\
			& \textit{Input} & \textit{Target} \\
			\hline         		
			AnswerPrompting & \makecell{Choice1: Oven Choice2: Refrigerator \\  Where do you put food to keep it cold?} & \makecell[l]{Refrigerator} \\
			\hline         		
			MaskedAnswerPrompting & \makecell{AnswerPrompting + \textcolor{brown}{\textbf{X}}} & \makecell[l]{\textcolor{brown}{\textbf{X}} Refrigerator} \\ 	
			\hline         		
			ChoicePrompting & \makecell{AnswerPrompting \\ + \textcolor{cyan}{the correct choice is}} & \makecell[l]{Choice2 Refrigerator} \\
			\hline         		
			MaskedChoicePrompting & \makecell{AnswerPrompting + \textcolor{brown}{\textbf{X}} \\ + \textcolor{cyan}{the correct choice is} \textcolor{blue}{\textbf{Y}}} & \makecell[l]{\textcolor{brown}{\textbf{X}} Refrigerator \\ \textcolor{blue}{\textbf{Y}} Choice2} \\
			\midrule
            \multicolumn{3}{c}{\textbf{QA Prompt-Tuning}} \\
            \midrule			
			PreAnswerPT & \makecell{\textcolor{red}{ $ \mathbf{P}_1 \mathbf{ P}_2  \dots  \mathbf{P}_n $} + AnswerPrompting} & \makecell[l]{Refrigerator} \\
			\hline
			PostAnswerPT & \makecell{AnswerPrompting + \textcolor{red}{ $ \mathbf{P}_1 \mathbf{ P}_2  \dots  \mathbf{P}_n $}} & \makecell[l]{Refrigerator} \\
			\hline
			PreMaskedAnswerPT & \makecell{\textcolor{red}{ $ \mathbf{P}_1 \mathbf{ P}_2  \dots  \mathbf{P}_n $} + AnswerPrompting + \textcolor{brown}{\textbf{X}}} & \makecell[l]{\textcolor{brown}{\textbf{X}} Refrigerator} \\
			\hline
			PostMaskedAnswerPT & \makecell{AnswerPrompting + \textcolor{red}{ $ \mathbf{P}_1 \mathbf{ P}_2  \dots  \mathbf{P}_n $} + \textcolor{brown}{\textbf{X}}} & \makecell[l]{\textcolor{brown}{\textbf{X}} Refrigerator} \\
			\hline
		\end{tabular}
		\caption{Templates for Formatting the Input and Output of an Encoder-Decoder Model during Fine-Tuning and Prompt-Tuning for  Question Answering Tasks.}
		\label{table:qa-methods}
	\end{table*}
	
%\end{document}

\subsubsection{Templates for Prompt-Tuning}
Table \ref{table:methods} also outlines the templates utilized in prompt tuning. These templates are similar to those used in fine-tuning but involve dedicated prompt tokens that are optimized during the training process. Notably, the position of these prompt tokens within the input serves as a parameter to be explored.

We investigate two primary templates. The first involves prepending the prompt tokens to the beginning of the input sequence (denoted with the "Pre" prefix). The second configuration entails appending the prompt tokens to the end of the input sequence, just before the placeholder or mask token (denoted with the "Post" prefix). This latter configuration bears a closer resemblance to the position of natural prompts discussed in the previous section.

\subsubsection{Templates for Fine-Tuning QA Tasks}
The reviewed templates were general and suitable for various tasks; however, task-specific templates can also be designed. In this section, we have created templates specifically for QA tasks to align them with a generative text-to-text approach and the objectives of the PLM. Table \ref{table:qa-methods} presents the templates designed for QA tasks:

\begin{itemize}
\item  \textbf{AnswerPrompting:}
This technique is similar to the prompting method, where the prompt is the question itself. We positioned the question at the end of the input to place it near the target answer, aligning the format more closely with the objective used in the pre-training phase. The model is trained to generate the text of the correct choice. Our hypothesis is that the model can utilize its training objective to predict the complement and select the correct choice. Similar to prompting, we expect this method to be suitable for longer target texts.

\item \textbf{MaskedAnswerPrompting:}
This method offers an alternative to AnswerPrompting, utilizing a denoising objective and masked tokens. Here, the model generates text associated with the correct choice for masked tokens specified in the input. Similar to MaskedPrompting, we anticipate this technique to be more suitable and perform better for questions with shorter answers.

\item  \textbf{ChoicePrompting:}
This technique resembles the AnswerPrompting method, where the prompt follows the question, asking for the correct choice. The model is trained to generate the text of the correct choice, as well as to specify the choice number itself.

\item \textbf{MaskedChoicePrompting:}
In this technique, the model generates both the correct choice number and its associated text for two masked tokens specified in the input. Similar to MaskedPrompting, we anticipate this method to be particularly effective and perform well for questions with shorter answers.
\end{itemize}

We could choose to ignore predicting the choice number and focus solely on the choice text. However, our experiments show that predicting both the choice number and its text results in slightly higher accuracy in certain models. This improvement is likely because predicting the choice number helps the model stay more focused on the given choices.

\subsubsection{Templates for Prompt-Tuning QA Tasks}
Table \ref{table:qa-methods} displays the templates specific to prompt tuning for QA tasks. These templates correspond to the aforementioned templates used in fine-tuning. However, as the prompts are fine-tuned, we explored the option of placing them at the start of each case or at the end of each case after the question.

In the following section, we delve into the integration of various models, pretraining objectives during the adaptation stage, and fine-tuning templates to assess their performance and elucidate the role of each component in achieving optimal results across diverse tasks. Our aim is to discern a recipe for enhancing performance across various tasks using both fine-tuning and prompt-tuning methods.

\section{Experimental Setup}
\subsection{Datasets}
In our experiments, we specifically chose tasks categorized as shown in Table \ref{table:dataset}. Within the scope of the first and second task categories, we drew upon relations sourced from the $\text{ATOMIC}^{20}_{20}$ dataset \citep{comet2020}. $\text{ATOMIC}^{20}_{20}$ is the extension of the ATOMIC dataset \citep{atomic}, serving as a commonsense knowledge base, encompassing 23 distinct types of commonsense relations. These types can be broadly categorized into three groups: 9 social-interaction commonsense relations, 7 commonsense relations related to physical entities, and 7 event-centered commonsense relations that pertain to situations surrounding specific events of interest.

The entries within this dataset adopt a $(\textbf{head, relation, tail})$ triplet format. These tuples can be structured as natural language sentences. For instance, the sentence "if \underline{PersonX cooks}, he intends \underline{to satisfy hunger}" illustrates the intention relation, denoted by \textbf{xIntent} in the dataset.  Notably, each head may correspond to multiple tails for each relation. These tails represent plausible or potential consequences of the given head event.

%For the classification tasks, we engaged with three specific assignments: MNLI and QNLI, both categorized under Natural Language Inferences, and SST2, which pertains to sentiment analysis.  These tasks are part of GLUE benchmark for Natural Language Understanding (NLU)\citep{glue}.We posit that sentiment analysis can be enhanced through prompting techniques, given its relatively straightforward nature. On the other hand, tasks related to Natural Language Inferences (NLI) necessitate a deeper understanding of sentence fragments and their complements.

For the question-answering (QA) tasks, we engaged with two specific assignments: CommonsenseQA \cite{commonsenseqa} and OpenBookQA \cite{openbook}. These tasks are part of the broader evaluation of QA systems, testing their ability to understand and reason about text. Specifically, we used the official split of CommonsenseQA and evaluated our methods on its official development split.

	\begin{table}[t]
		\centering

		\begin{tabular}{lcccc}
			\toprule
			\textbf{Dataset} & \textbf{Size} & \textbf{Q. Length} (avg.) & \textbf{Choice Length} (avg.) & \textbf{Q. Ending with '?'} (\%) \\
			\midrule
			CSQA (train) & 9.7K & 13.25 & 1.52 & 99.31 \\
			CSQA (test) & 1.2K & 13.43 & 1.51 & 99.04 \\
			\midrule
			OBQA (train) & 5K & 10.71 & 2.78 & 35.69 \\
			OBQA (test) & 500 & 10.30 & 3.06 & 35.80 \\
			\bottomrule
		\end{tabular}
			\caption{Statistics of the QA datasets, with lengths based on average word counts.}
		\label{table:qa-datasets}
	\end{table}

CommonsenseQA is designed to evaluate a system's proficiency in comprehending commonsense knowledge and reasoning, necessitating an understanding that extends beyond the given text. In contrast, OpenBookQA evaluates the system's capability to apply fundamental scientific knowledge, often requiring multi-step reasoning to reach the correct answer. Additionally, these datasets differ in the structure of questions and the length of answer choices, aspects of interest in this research. A summary of these two datasets is presented in Table \ref{table:qa-datasets}.

For our unsupervised training stage, we relied on the Open Mind Common Sense (OMCS) corpus, a large-scale collection of commonsense knowledge consisting of over 700,000 sentences in natural language \cite{omcs}. The sentences in the OMCS corpus are assigned relevance scores by its developers. We ranked the sentences based on these scores and selected the top 8,000 sentences to train the models in an unsupervised manner. These selected sentences were then split using our proposed method in Section \ref{sec:splitter} to prepare the training data. We used selected tasks from the Atomic dataset to provide 5,000 samples to train the task classifier and sentence splitter models. The tasks and their categories are provided in the Appendix (\ref{table:filling_mapping}).

%For QA tasks we used CommonsenseQA (Multi-choice questions), CommonsenseQA-2 (yes/no questions), which are questions about ??? \cite{comqa}, PIQA \cite{piqa}, questions, multi-choice questions about physical objects and social-qa multi-choice questions about ???

We employed the templates outlined in the appendix (Table \ref{table:tuples}) to structure and format the relations within the first and second categories.

\subsection{Model}
Our method is primarily designed for text-to-text encoder-decoder models. For our experiments, we chose to focus specifically on the T5 model \citep{t5} due to its suitability for our purposes and the availability of various versions that serve as the basis for our comparisons. The T5 model has been successfully employed for QA and KB completion tasks and has outperformed other models in these domains \citep{comet2020, analyzingCS, rainbow, unified}.
\begin{table*}[tbh!]
	\centering
	\begin{tabular}{|l | l | c|}
		\hline
		\textbf{Model} & \textbf{Pre-training Objective} & \textbf{Data Sets}   \\ 
		\hline
		T5-v1   &   Noise Replace Spans  & C4   \\ 
		\hline
		T5-lm   &  \makecell[l]{Noise Replace Spans \\ + Language Modeling}  & \makecell[c]{C4 \\ + 100K Steps on the LM Objective } \\
		\hline		
		T5-base   &  \makecell[l]{Noise Replace Spans  \\ + Language Modeling \\ +  Supervised Text-to-Text}  & \makecell[c]{C4  + WikiDPR  \\ Various Supervised Tasks} \\
		\hline				
	\end{tabular}
	\caption{Different Versions of the T5 Model Based on the Pre-training Objective.}
	\label{table:models}
\end{table*}

The T5 model functions as an encoder-decoder model and has undergone pre-training using both unsupervised and supervised methods, each with distinct pre-training objectives. Multiple versions of the T5 model have been released by its creators \citep{t5}. The models utilized in our experiments are displayed in Table \ref{table:models}. These versions primarily diverge in their pre-training objectives and certain architectural aspects. 

The T5-v1 model was exclusively pre-trained on the C4 dataset \footnote{\url{http://www.tensorflow.org/datasets/catalog/c4}} employing an unsupervised denoising objective. The T5-lm model, a derivation of T5-v1, underwent an additional 100k training steps focused on the prefix language model (LM) objective. This supplementary training phase aimed to augment its adaptability for prompt tuning purposes.

The third model is the T5-base model, a widely adopted choice for a variety of downstream tasks. Preceding its application in these tasks, the T5-base model underwent a pre-training phase characterized by a multitasking methodology. This approach integrated a blend of both supervised and unsupervised tasks. As with other models, during its unsupervised training, the T5-base model utilized the denoising objective by strategically replacing spans. T5-large is its counterpart, boasting increased capacity and enhanced performance compared to T5-base.

\subsection{Implementation Details} 
Across all models, we employ the Ada Factor optimizer \cite{ada} with a consistent learning rate of 0.0001 throughout the fine-tuning process. The mini-batch size is set at 8 for T5-base and 4 for T5-large, and the model undergoes training for a total of 3 epochs for fine-tuning. 

In the context of prompt tuning, we established the length of prompt tokens to be exactly 10 tokens for KB completion tasks and 20 tokens for QA tasks. The batch size was set to 32 for KB completion tasks and 8 for QA tasks. The training epochs were 6 for QA tasks and 12 for KB completion tasks. 

Nonetheless, directly optimizing the embeddings of prompt tokens can introduce instability, potentially causing the optimizer to converge to local minima, as indicated by prior studies \citep{gptund, prefix}. Similar to these studies, we adopted a multilayer perceptron neural network as the prompt encoder to optimize the prompt embeddings.

\begin{align}
	\text{h}_i &= \textbf{MLP}(\text{h}_i)
\end{align}

The prompt encoder operates on each prompt token's embedding individually, resulting in a transformed embedding $\mathbf{h}_i$. This modified embedding, combined with the input embeddings, is then fed into the language model. We adopted a learning rate of 0.08 for the prompt encoder. 

%For our prompt-tuning experiments, we opted to utilize a set of $n=100$ examples per relation, aiming to ensure meaningful and robust outcomes. Simultaneously, we extended the training epochs to a count of 20, thereby affording more opportunities for the model to converge and learn effectively.

\subsection{Evaluation}
For QA tasks, we used accuracy as the evaluation metric to measure the correctness of the model's predictions against the ground truth labels.

For generation tasks, the generated tails corresponding to a given head and relation can be evaluated in terms of plausibility. This evaluation can be conducted using both automated metrics, such as ROUGE \citep{rouge} and BERT score \citep{bertscore}, as well as human verification. In this paper, our evaluation primarily relies on automated metrics. However, we also manually reviewed the generated results to ensure their validity as coherent and meaningful text. The automated evaluations were conducted on 100 unique heads, each associated with a minimum of three target outputs, resulting in an evaluation dataset of approximately 300 instances per relation.

For each distinct head,  a score is computed by comparing the generated output with the corresponding tails associated with that head. Subsequently, the highest score is attributed to the particular instance. The ultimate score is derived as the average of the scores assigned to all instances. 

For QA tasks, we utilized the test split of each dataset.

\begin{figure}[h!]
	\centering
	\includegraphics[width=\textwidth]{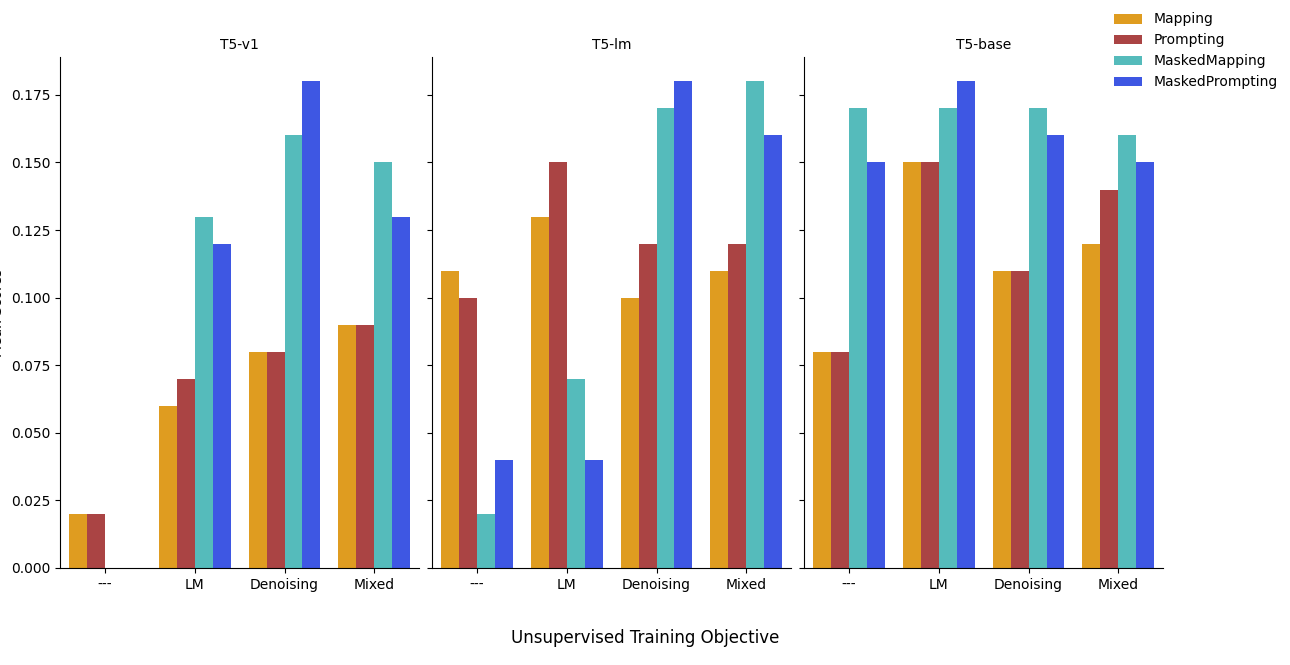}
	\caption{ ROUGE scores evaluating the tasks in Mask-Filling category across various models with distinct pretraining and fine-tuning objectives. Each task is evaluated with a sample size of $n=$ 30, over 3 epochs. "--" indicates direct fine-tuning with no adaptation stage.}
	\label{fig:ft-filling}
\end{figure}

\section{Results and Discussion}
In the following sections, we present the results for each model across different task categories for KB completion tasks. We begin by discussing the results for the fine-tuning approach, followed by the results for the prompt-tuning approach. Based on these findings, we refine our models and techniques for QA tasks and compare our results with related works.

\subsection{Fine-Tuning Methods}

\subsubsection{Mask-Filling Performance} 
The results for various fine-tuning approaches within the Masked-Filling category are showcased in Figure \ref{fig:ft-filling}. These evaluations were conducted using $n=30$ examples per task over three epochs. The scores for each bar represent the average performance across the tasks within this category. The results presented here are based on ROUGE scores; the corresponding BERT scores can be found in Appendix \ref{sec:appendix-bert}. In our experiments, the ROUGE, BLEURT, and BERT scores are highly correlated and exhibit similar trends, so we focus on the ROUGE scores here (refer to Figure \ref{fig:rg-bert-bleu} in Appendix).

Figure \ref{fig:ft-filling} illustrates three distinct sections for the three models. Each section is further divided into four parts: the first part represents the model's performance when directly fine-tuned, while the remaining three parts show the model's performance after undergoing our proposed unsupervised training via three different objectives (LM, Denoising, and Mixed). 

As indicated in the figure, extending the unsupervised training on related natural sentences is highly effective in improving performance for all models. However, the different objectives used during this phase affect performance variably depending on the model and the fine-tuning template used to format the task-specific data. Consistently, templates using masked tokens (\textbf{MaskedMapping} and \textbf{MaskedPrompting}) outperform the mapping methods (Mapping and Prompting). This advantage likely arises from the nature of the tasks within this category, which can be effectively addressed by finding a suitable word for the masked token.

In the case of T5-v1 and T5-LM, the highest performance is achieved when the adaptation stage is conducted using the \textbf{Denoising} objective, and the fine-tuning stage employs MaskedMapping or MaskedPrompting templates. It's noteworthy that these templates were not effective when the LM objective was used in the adaptation stage. The T5-v1 model exhibits inferior performance in the absence of unsupervised training. 

In the case of T5-LM, whether the model is directly applied or undergoes unsupervised training using the LM objective, the performance is higher when corresponding templates are used in the fine-tuning stage (namely Mapping and Prompting). However, Denoising or Mixed objectives together with Masking templates are more effective due to the nature of the tasks. Notably, these objectives positively affect MaskedMapping. An advantage of MaskedMapping over MaskedPrompting is the elimination of the need for prompt engineering.

Given the relatively lower results of T5-lm using the Masked templates with the LM objective, it is plausible that the inclusion of an additional 100K pre-training steps dedicated to the LM objective in T5-LM pre-training could have led to a partial reduction in the proficiency gained through pre-training with Denoising objective. This underscores the importance of employing Denoising objectives in the adaptation stage before using the associated templates. 

%TODO Epochs
%Furthermore, in the context of T5-base, the extension of training epochs led to a reduction in the performance of MaskedPrompting, despite it being the best-performing method with 5 epochs. This phenomenon could be attributed to overfitting.

In the case of T5-base, it achieves high scores when using Masked templates in fine-tuning stages. Its direct use exhibits comparable performance with the results after unsupervised training. However, employing the LM objective can further boost the performance of the related templates (Mapping and Prompting), and overall, this objective leads to higher results when Masked templates are utilized.
 
The T5-base model underwent pre-training, incorporating a blend of supervised and unsupervised objectives in a multitasking fashion. This distinct pre-training strategy empowers the model to deliver robust performance across both formatting methods, without compromising the efficacy of either approach. Notably, for the Mask-Filling tasks, the Denoising objective appears to be particularly efficacious.

%As seen from Table \ref{table:top-ft-filling}, the performance of different tasks is also impacted by the phrasing utilized to formulate tuples in natural language. For instance, the success of the HasProperty and CapableOf relations across all models, especially the T5-lm model, does not align with the achievements observed for ObjectUse or AtLocation. This variation can be attributed to the employment of the less common phrase "X has property of Y" for the HasProperty relation, as opposed to the more prevalent phrasing like "X is used for Y" or "X is located at Y" found in the ObjectUse and AtLocation relations. The higher quality of these relationships is also evident from the BERT scores presented  in Appendix (Table \ref{table:res-group1-bert}).

%	\caption{ ROUGE scores assessing the quality of tail generation across different models using diverse fine-tuning techniques. Each task is evaluated with a sample size of $n=$ 30, over 5 epochs, with average scores reported for both 5 and 15 epochs.}
%	\label{table:ft-pcl-results}
	
\begin{figure}[t]
	\centering
	\includegraphics[width=\textwidth]{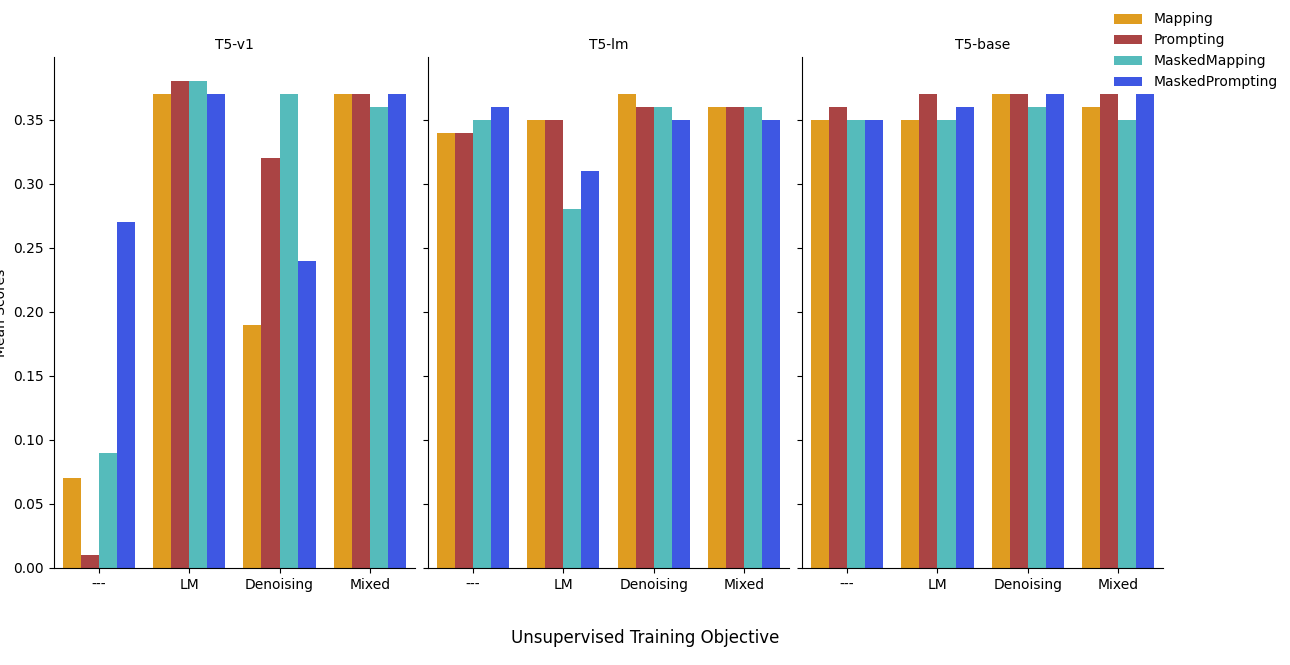}
	\caption{ROUGE scores evaluating the tasks in Map-Phrasal category across various models with distinct pretraining and fine-tuning objectives. Each task is evaluated with a sample size of $n=$ 30, over 3 epochs}
	\label{fig:ft-map}
\end{figure}

\subsubsection{Map-Phrasal Performance}
Figure \ref{fig:ft-map} displays the performance of models across the Map-Phrasal category. In this context, a trend similar to Mask-Filling is observed across different models. However, the performance gap between different methods within a model is relatively smaller.
%TODO refine or omit
%Also note that the automatic evaluation within this category is difficult.

The effect of the unsupervised training stage is most obvious and influential for the T5-v1 model. The LM objective in this training can noticeably increase the performance of the model across all four templates. Note that in the case of T5-LM, the Denoising objective is slightly more effective than the LM objective. This suggests that the objective in the adaptation stage is more effective when it complements the main objective in the pre-training phase. The Mixed objective consistently shows good performance across all models.

Unlike the Mask-Filling category, where the denoising objective combined with masked templates outperformed, here the language modeling (LM) and mixed objectives combined with mapping templates prove more effective. This could be attributed to the nature of these tasks, which require longer outputs and an understanding of relationships that can be generalized to new situations. In any case, the combination of objectives, either at the pre-training stage (as in T5-base) or as a complementary stage after pre-training, is beneficial for various tasks.

%This trend aligns with the results presented earlier, highlighting the relationship between task types and pre-training objectives in influencing task performance. In general, the denoising objective is well-suited for tasks involving sentence completion with familiar constructs that the model encountered during pre-training. Conversely, the LM objective is more effective for learning from examples and generalizing to unfamiliar data.

\begin{figure}[t]
	\centering
	\includegraphics[width=\textwidth]{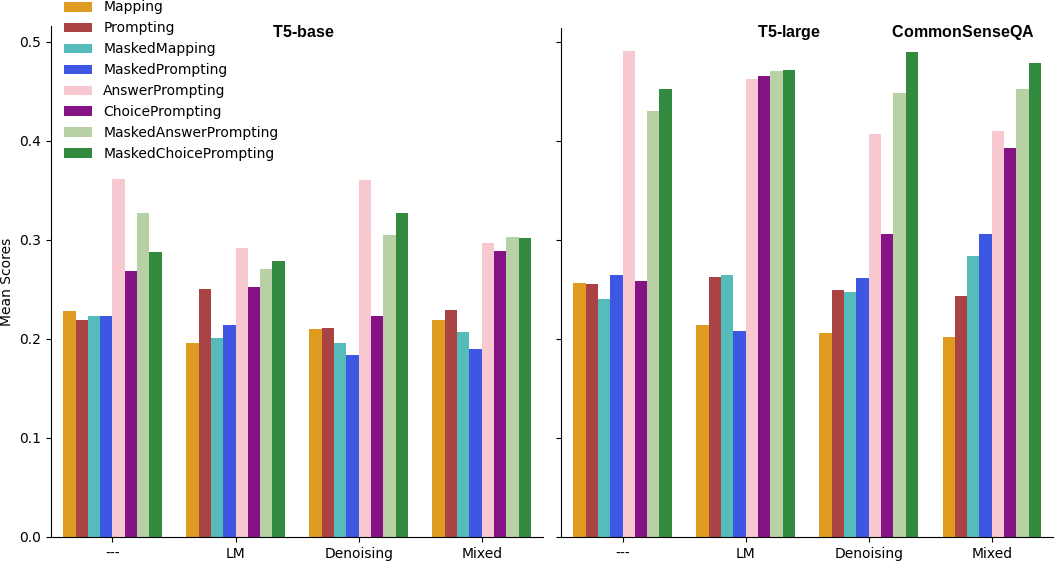}
	\includegraphics[width=\textwidth]{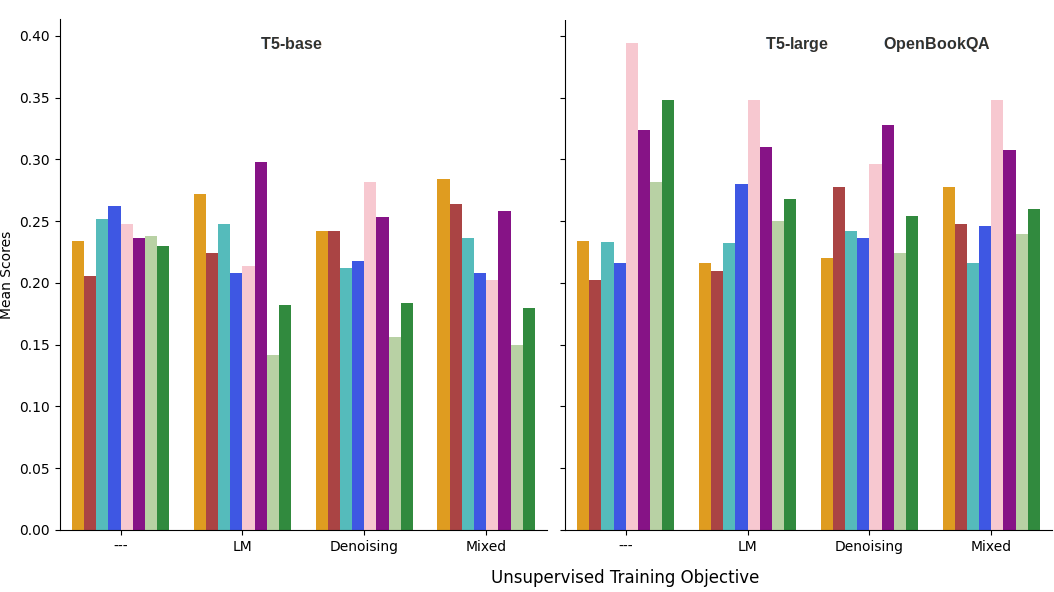}
	\caption{Accuracy scores evaluating tasks in the question answering category: CommonsenseQA (top) and OpenBookQA (bottom). The performance of T5-base and T5-large models with different pretraining and fine-tuning objectives is presented. Each task is assessed with a sample size of $n=30$, over 3 epochs.}
	\label{fig:ft-qa}
\end{figure}

\subsubsection{Question Answering Performance} \label{sec:qa}
Based on our observations of the performance of different models across two task categories, we chose T5-base for our QA experiments due to its strong performance in both objectives. To examine the effect of scale, we also included T5-large. 

For QA tasks, we assessed our general templates (Table \ref{table:methods}) alongside those specifically designed for QA tasks (Table \ref{table:qa-methods}). Figure \ref{fig:ft-qa} presents the performance of different templates on T5-base and T5-large models, after adaptation phase with various objectives, on CommonsenseQA and OpenBookQA tasks.

The QA-specific templates substantially outperform both our general templates and the conventional sequence-to-sequence fine-tuning baseline. Performance varies depending on the pre-training stage and the task characteristics. The performance boost is more pronounced with T5-large, demonstrating the effectiveness of the proposed strategies in distilling knowledge from larger models.

For CommonsenseQA, Masked templates (MaskedAnswerPrompting and MaskedChoicePrompting) outperform the other methods, particularly when the Denoising objective is used in the adaptation phase. However, for OpenBookQA, AnswerPrompting and ChoicePrompting perform better, while Masked templates do not work as well. Referencing Table \ref{table:qa-datasets}, the different characteristics of these tasks can explain this difference. Firstly, the answers in CommonsenseQA are generally shorter and simpler. As noted, Masked templates are effective for such questions. Moreover, over 99\% of CommonsenseQA questions end with a question mark, providing consistent prompting for the masked answer. Additionally, we trained the models using the OMCS corpus, which aligns well with common sense questions. In contrast, OpenBookQA questions are more complex, requiring longer answers, and most are not in question format but rather open-ended sentences or phrases. For this format, the LM objective performs better, as seen with AnswerPrompting. Examples of questions from these datasets can be found in the Appendix \ref{apx:ds-examples}. 

\begin{figure}[h!]
	\centering
	\includegraphics[width=\linewidth]{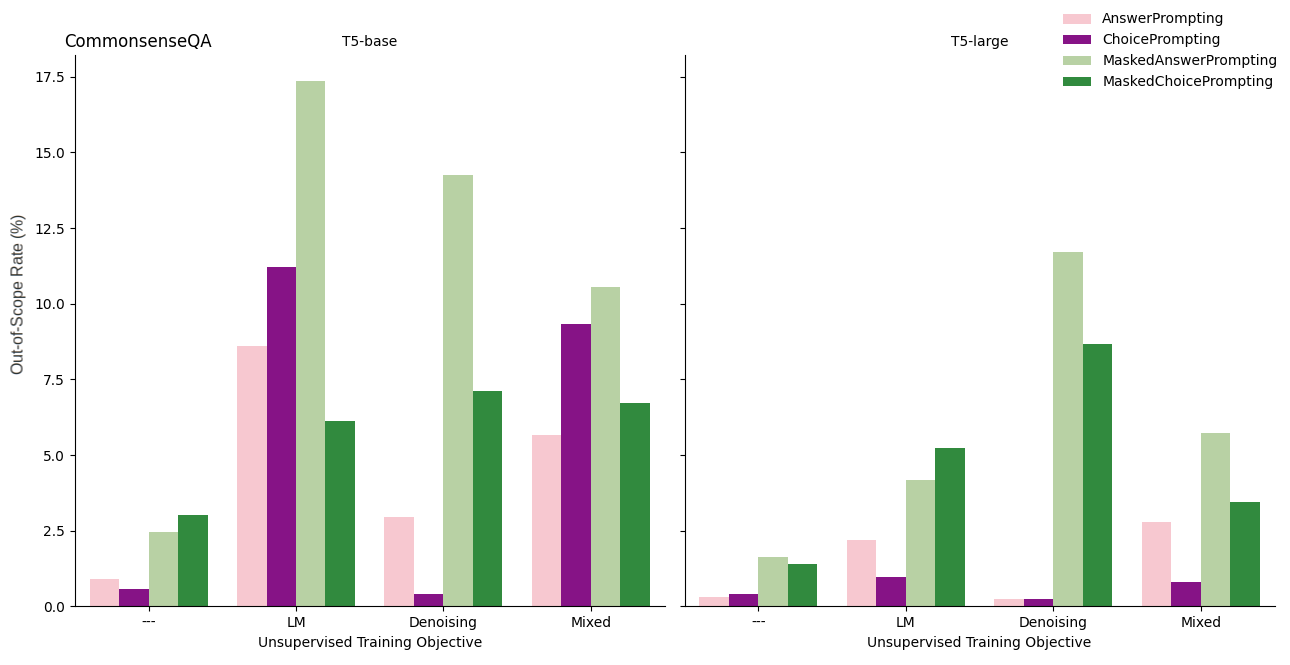}
	\includegraphics[width=\linewidth]{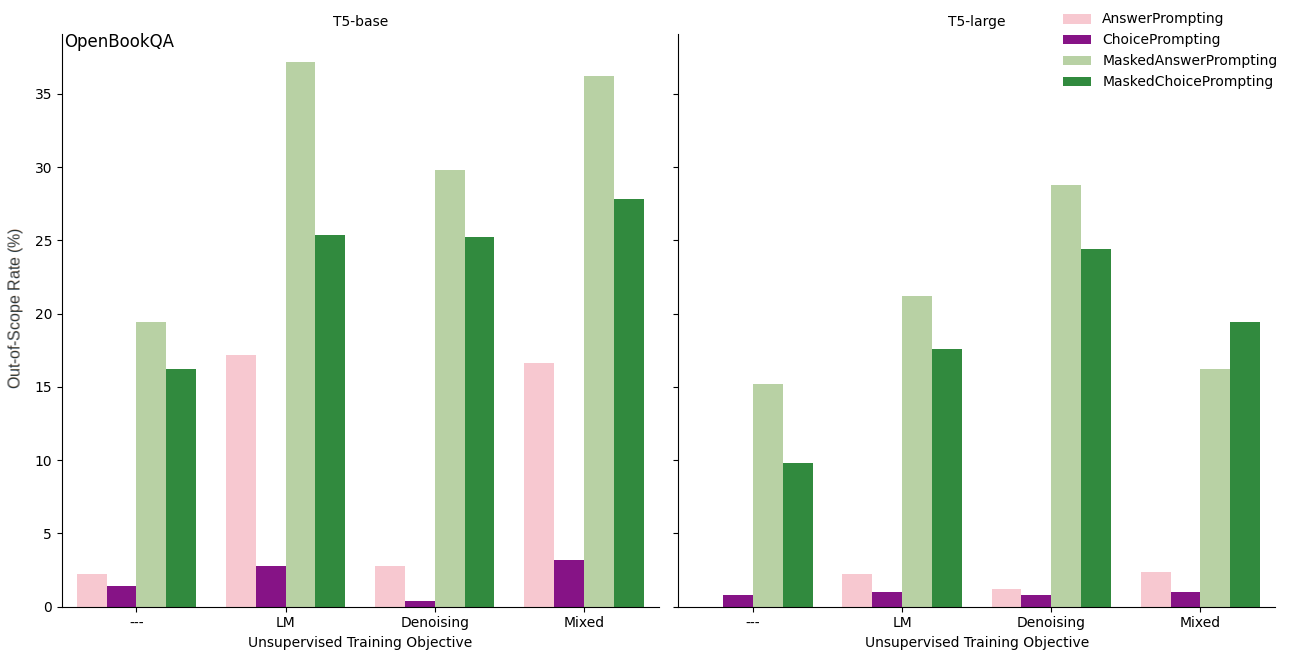}
	\caption{The percentage of predictions that are outside the given choices of Multiple Choice QA for different methods across the CommonsenseQA dataset (top) and OpenBookQA dataset (bottom).}
	\label{fig:ft-cs-qa-out}
\end{figure}

\subsubsection{Analysis of Prediction Errors} \label{sec:out-of-scope}
Upon examining the model predictions, we noticed a trend across various methods: they may generate answers that are not listed among the given choices, particularly in few-shot scenarios. However, these out-of-scope responses may still be valid, as common-sense questions often allow for answers beyond the provided options. For instance, consider the following example from the predictions of the model using MaskAnswerPrompting template:

\begin{quote}
	Choices: 1. \textcolor{teal}{\textbf{build trust}}, 2. hurry home, 3. ignore people, 4. believe in god \\
	Question: \textbf{What do people do when networking?} \textcolor{brown}{X} \\
	Prediction: \textcolor{brown}{X} \textcolor{blue}{talk to each other}
\end{quote}

The prediction "talk to each other" is a plausible response that isn't among the given choices but still makes sense in the context of networking. Additionally, in some cases, the question itself may contain a fallacy, or the format of the predicted answer might differ from the correct choice, such as "getting wet" versus "get wet." You can find further examples illustrating these variations in Table \ref{table:qa-predictions} in the Appendix.

We computed the percentage of out-of-scope predictions for different methods across two datasets, as illustrated in \ref{fig:ft-cs-qa-out}. Notably, this tendency is significantly lower in prompting templates that do not employ masked tokens (AnswerPrompting, ChoicePrompting). We attribute this difference to their respective objectives: prompting methods are aligned with the language modeling objective, which involves completing a broken sentence with a fragment, thus emphasizing the relevance of the entire context. Conversely, masked templates align with the denoising objective, typically necessitating the completion of one or more missing words, which directs attention toward nearby words. Consequently, prompting methods devote more attention to the question and given choices, thereby leveraging information from supervised examples during few-shot training.

Despite this difference, which can decrease the accuracy of masked templates, their overall performance remains high and comparable to prompting methods. Interestingly, when the model uses the Denoising objective during the adaptation phase, the rate of out-of-scope answers increases, yet the accuracy is still substantially higher than the baseline (no adaptation). This indicates that while this objective might negatively impact accuracy (based on matching with given choices), it can still be useful in identifying the correct answer in other examples.

\subsubsection{Effect of Choice-Prompting Methods}
To mitigate the rate of out-of-scope answers while still leveraging the benefits of denoising and prompting methods, we introduced ChoicePrompting and MaskedChoicePrompting, as detailed in Table \ref{table:qa-methods}. These approaches require predicting the choice number alongside the answer, aiming to compel the model to focus more on the provided choices. As depicted in Figure \ref{fig:ft-cs-qa-out}, this strategy effectively reduces the out-of-scope rate for both AnswerPrompting and MaskedAnswerPrompting, resulting in the highest performance when the model is adapted with the denoising objective using MaskedAnswerPrompting. However, in the case of AnswerPrompting without adaptation, the negative impact of this approach, compared to unadorned answers, which are more natural, outweighs its benefits.

\begin{figure}[t]
	\centering
	\includegraphics[width=1\linewidth]{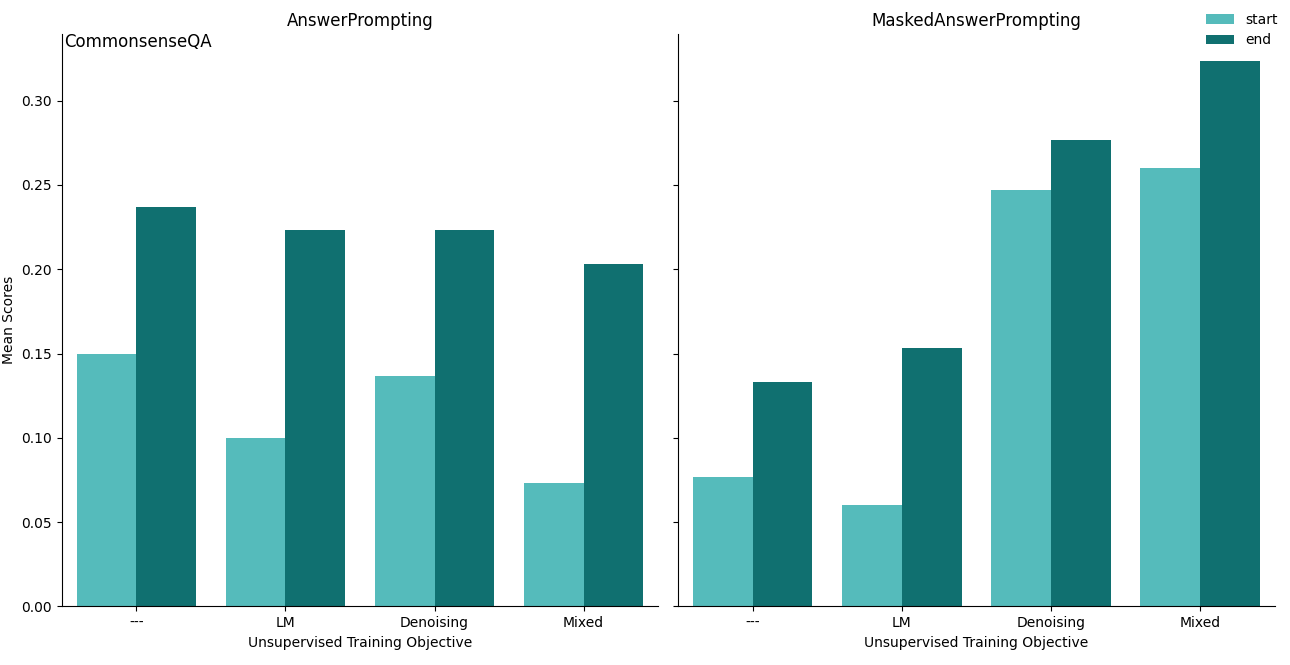}
	\includegraphics[width=0.49\linewidth]{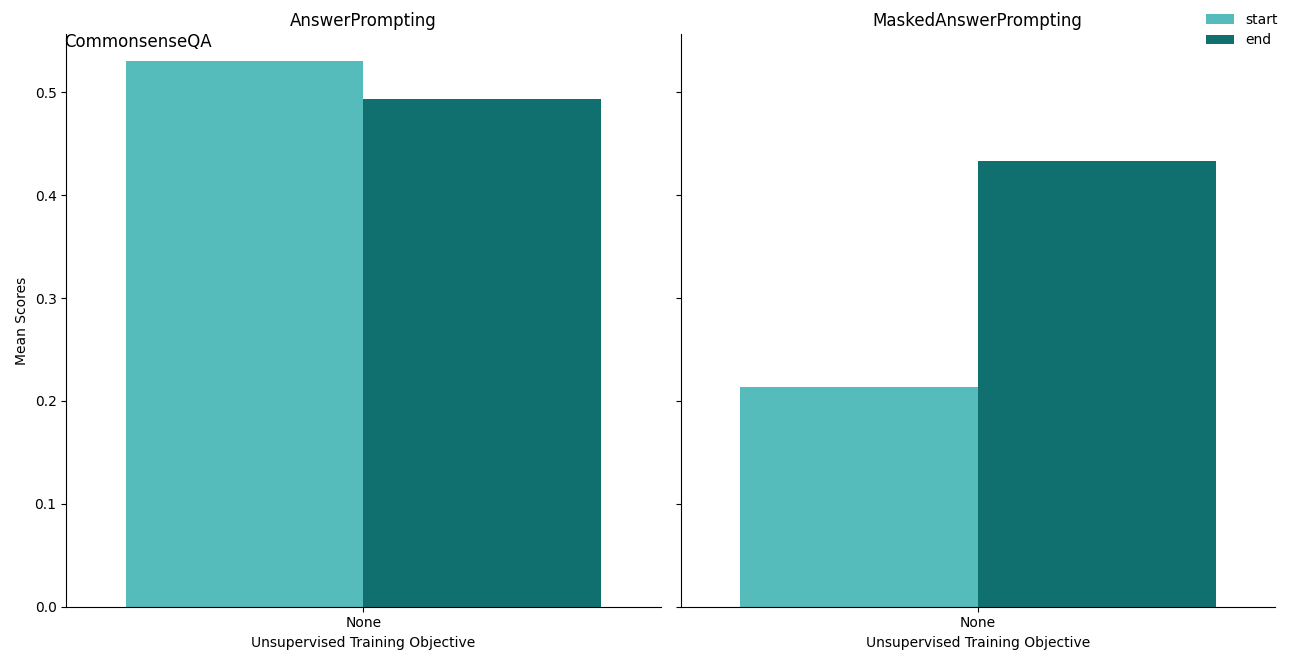}
	\includegraphics[width=0.49\linewidth]{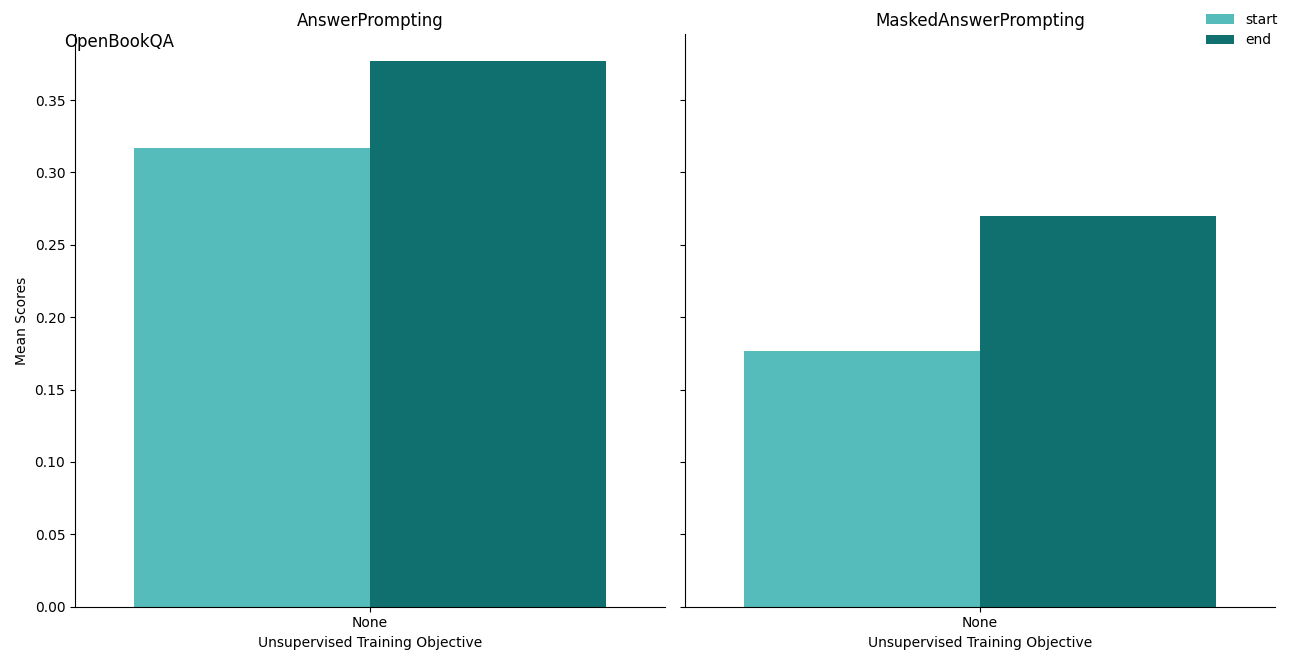}
	\caption{Effect of question position (start vs. end of input) in AnswerPrompting and MaskedAnswerPrompting for T5-lm (top) across different unsupervised objectives for the CommonsenseQA task. Bottom: Effect of question position for the T5-large model on CommonsenseQA (left) and OpenbookQA (right).}
	\label{fig:qpos}
\end{figure}

\subsubsection{Impact of Question Position in Input}
In designing our QA templates, we hypothesized that positioning the question at the end of the input, adjacent to the target answer, could improve performance by being more aligned with the LM and denoising objectives. Figure \ref{fig:qpos} compares the performance of T5-lm with the question placed at the beginning and at the end of the input across different adaptation objectives over CommonsenseQA. It is clearly evident that placing the question at the end is effective for both AnswerPrompting and MaskedAnswerPrompting, which correspond to the LM and denoising objectives, respectively.

The figure also shows the effect of question position using T5-large across CommonsenseQA and OpenBookQA. For MaskedAnswerPrompting, the trend observed for T5-lm holds, with placing the question at the end being more effective. However, for CommonsenseQA using AnswerPrompting, placing the question at the beginning of the input is slightly better. It is important to note that T5-base and T5-large undergo supervised training on various tasks, including QA tasks, and the formats used in these tasks could bias the model toward a specific format. Nevertheless, for OpenBookQA, where the questions act more as prompts (refer to \ref{table:openbook}), the effect of placing the question at the end is more pronounced. In general, prompting templates compared to masked methods are less sensitive to the position of the question in this case.

\begin{figure}[t]
	\centering
	\includegraphics[width=\textwidth]{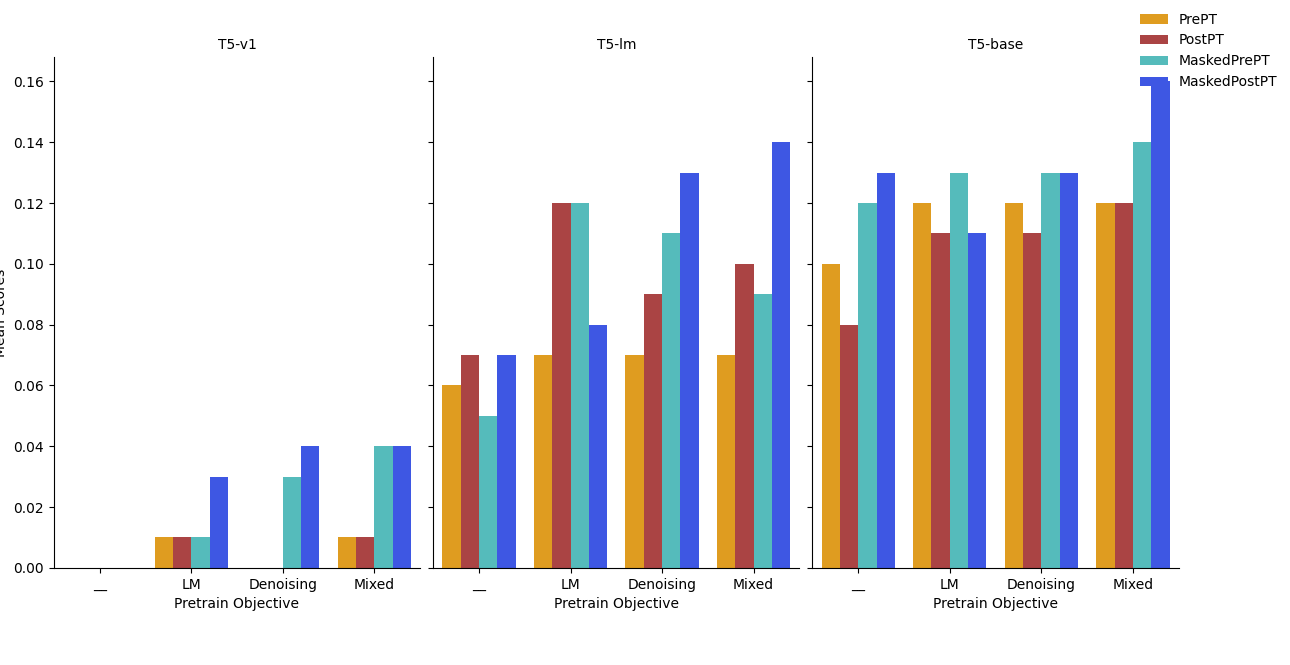}
	\caption{ROUGE scores evaluating the tasks in Mask-Filling category across various models with distinct pretraining and prompt-tuning objectives. Each task is evaluated with a sample size of $n=$ 30, over 3 epochs}
	\label{fig:pt-filling}
\end{figure}

\subsection{Prompt Tuning Results}
Similar to the Fine-Tuning section, Figures \ref{fig:pt-filling} and \ref{fig:pt-map} showcase the performance of all models across three mentioned categories of tasks.
\subsubsection{Mask-Filling Performance}
According to Figure \ref{fig:pt-filling}, the \textbf{MaskedPostPT} approach, which is roughly analogous to MaskedPrompting in the fine-tuning methods, outperforms the alternative approaches for Mask-Filling tasks across all models, particularly when the model undergoes adaptation phase using Denoising or \textbf{Mixed} objectives. In this setup, the combination of prompt tokens and masked tokens, placed in close proximity, seems to contribute to improved performance for tasks within this category. Notably, in this case, we do not use natural language prompts for the relations and instead allow the soft prompt tokens to serve the role of a natural prompt. Meanwhile, PrePT and MaskedPrePT use natural prompts for the relations. This suggests that soft prompts could function similarly to natural prompts and even outperform them for certain relations.

\begin{figure}[t]
	\centering
	\includegraphics[width=\textwidth]{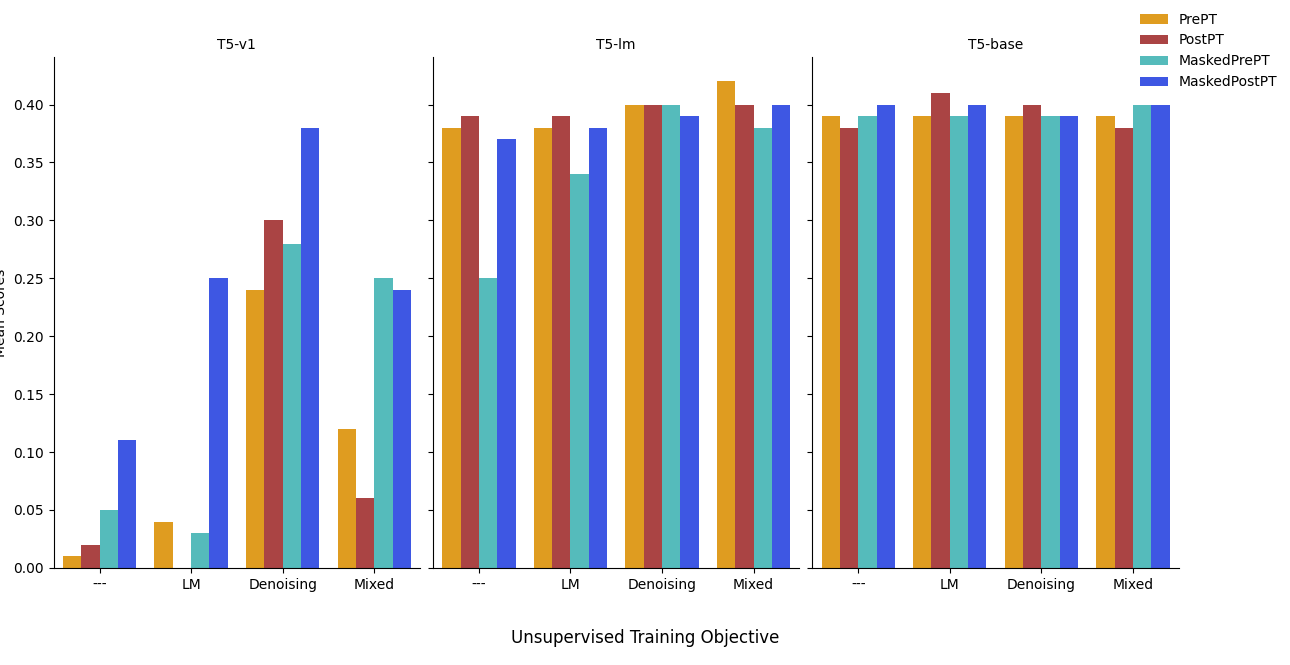}
	\caption{ROUGE scores evaluating the tasks in Map-Phrasal category across various models with distinct pretraining and prompt-tuning objectives. Each task is evaluated with a sample size of $n=$ 30, over 3 epochs}
	\label{fig:pt-map}
\end{figure}

An important observation here is that Prompt-Tuning is largely impractical with T5-v1 within a few-shot setting and Mapping methods (PrePT and PostPT). This phenomenon can be observed across other task categories as well, as indicated in Figures \ref{fig:pt-map}. We posit that the absence of pre-training in the LM objective renders the model ill-suited for prompt-tuning, wherein prompt tokens are intended to be optimized to serve as effective natural language prompts in a continuous space. The effect of the LM objective on PostPT and MaskedPrePT, which use natural prompts, is evident in the T5-LM and T5-base cases.

\subsubsection{Map-Phrasal Performance}
According to Figure \ref{fig:pt-map}, in the context of Map-Phrasal tasks, the difference between different templates is negligible, with PrePT and PostPT slightly outperforming the others. These approaches closely align with the Mapping and Prompting techniques in the fine-tuning paradigm, fitting well with the LM objective on which these models were trained.

The relations in this category require a deeper understanding of input examples and the ability to generalize them to new cases. The common and initial prompt at the beginning of examples likely facilitates making such connections. Moreover, the head and tail in this group can roughly complement each other without requiring a natural prompt (e.g., PersonX cook | to satisfy hunger). Additionally, automatic evaluation of these relations is more challenging than for Masked-Filling tasks.

%However, it's noteworthy that the performance of \textbf{MaskedPostPT} is comparable to these aforementioned methods, particularly after Denoising or Mixed objectives in adaptation stage. 
%Furthermore, across KB completion tasks encompassing the Mask-Filling and Map-Phrasal categories, MaskedPostPT consistently exhibits high performance, similar to MaskedMapping in the fine-tuning case. Thus, it can be considered a favorable method for tackling these specific tasks in the context of prompt-tuning. 
%This suggests that soft prompts could function similarly to natural prompts, rendering these two methods analogous.

\begin{figure}[t]
	\centering
	\includegraphics[width=0.8\textwidth]{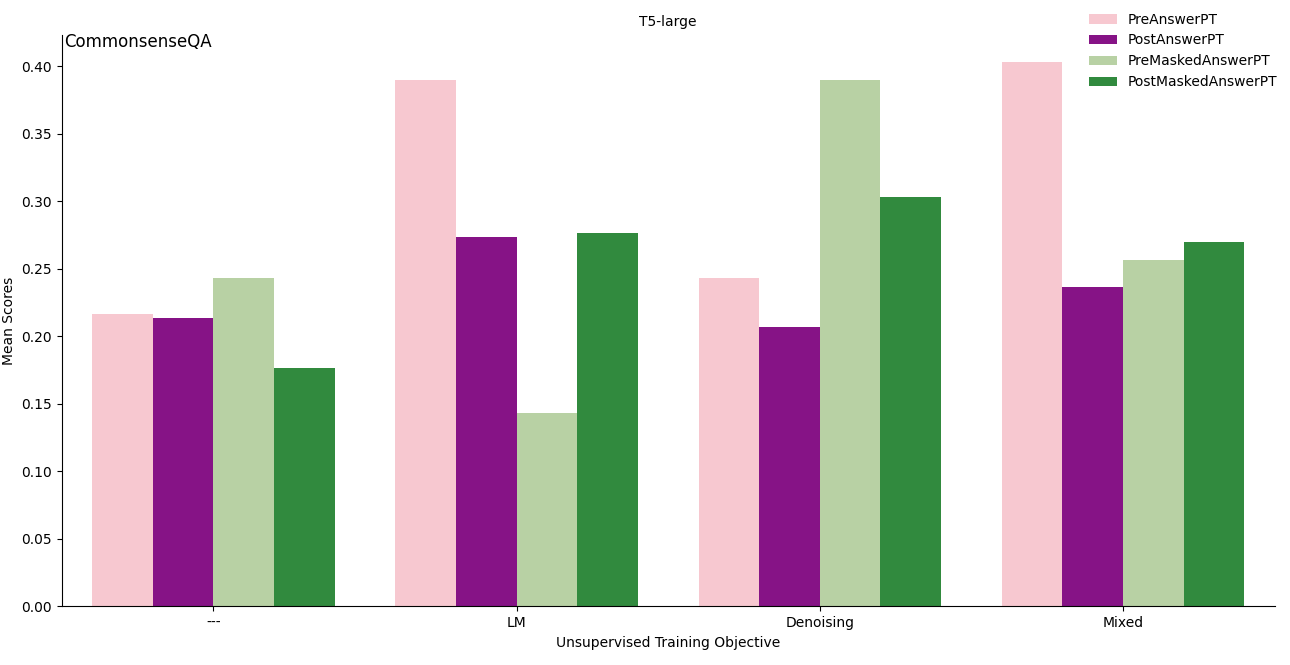}
	\caption{Accuracy scores evaluating CommonsenseQA across various models with distinct pretraining objectives and prompt-tuning methods.  It was evaluated with a sample size of $n=$ 90, over 6 epochs}
	\label{fig:pt-qa}
\end{figure}

\subsubsection{Question Answering Performance}
We also examined our designed templates specific to the Question Answering task, outlined in Table \ref{table:qa-methods}. These templates primarily correspond to AnswerPrompting and MaskedAnswerPrompting. For each method, there are two options: placing the prompt at the beginning of the input or after the question.

Figure \ref{fig:pt-qa} shows the results of prompt-tuning using T5-large on CommonsenseQA. We focused on CommonsenseQA since the training data during the adaptation phase is more aligned with this task. As observed, methods that place the prompts at the beginning of the input (PreAnswerPT and PreMaskedAnswerPT) outperform the other configurations, particularly when the adaptation phase objective matches these methods (LM and Denoising, respectively). Unlike Mask-Filling tasks, where MaskedPostPT performed better by placing the prompt tokens after the head to generate the tail, here placing the prompt before the question works better. Note that the question itself already functions as a prompt. Additionally, in the case of Task-Filling, the relation was constant across different samples (e.g., AtLocation), while for Question Answering, each question requires its own specific prompt.

\section{Comparison with Related Works}
In this section, we compare our results with related works in two categories: KB completion and question answering tasks. Specifically, we focus on QA tasks as they serve as benchmarks for numerous studies in the literature.

	\begin{table}[htbp]
		\renewcommand{\arraystretch}{1.3}
		\centering
		\begin{tabular}{lcccccc}
			\toprule
			\textbf{Task Category} & \textbf{Method} & \textbf{Adaptation} & \textbf{Template} & \textbf{ROUGE} & \textbf{BERT Score} \\
			\midrule
			\multirow{4}{*}{Mask-Filling} & \multirow{2}{*}{COMET \cite{analyzingCS}} & -- & Mapping & 9.02 & 43.74 \\
			& & -- & Prompting & 12.31 & 44.45 \\ 
			\cline{2-6}
			& \multirow{4}{*}{MTO (ours)} & -- & MaskedPrompting & 16.42 & 47.30 \\ 
			& & Denoising & Prompting & 13.89 & 46.47 \\ 
			& & Denoising & MaskedPrompting & 19.02 & 48.49 \\
			& & \textbf{Mixed} & \textbf{MaskedPrompting} & \textbf{19.92} & \textbf{49.54} \\
			\midrule
			\multirow{4}{*}{Map-Phrasal}  
			& \multirow{2}{*}{COMET}  & -- & Mapping & 39.02 & 52.89 \\
			& & -- & Prompting & 40.50 & 52.73 \\
			\cline{2-6}
			& \multirow{4}{*}{MTO (ours)} & -- & MaskedPrompting & 39.94 & 52.85  \\
			& & Denoising & MaskedPrompting & 41.8 & 53.63 \\
			& & \textbf{LM} & \textbf{Mapping} & \textbf{42.81 }& \textbf{56.31} \\
			& & Mixed & Mapping & 41.34 & 54.56 \\
			\bottomrule						
		\end{tabular}
		\caption{Comparison of our proposed models' scores with the baseline (Comet \cite{analyzingCS}) in few-shot settings, using $n=30$ samples on selected tasks from the $ATOMIC^{20}_{20}$ dataset. Additional score details for each taask are available in Appendix \ref{sec:scores}.}
		\label{table:results-kb}
	\end{table}

\subsection{Knowlege Completion}
\subsubsection{Fine Tuning}
In the case of KB completion on the ATOMIC dataset, significant work has been done by \cite{comet} and later by \cite{analyzingCS}. Both studies fine-tune a T5 model on examples from the ATOMIC dataset \cite{atomic} and apply it to new tuples. The latter also investigates techniques such as prompting in few-shot settings, which aligns with the Prompting method in our proposed approach. 

We demonstrated that aligning task templates during fine-tuning with the pre-training objective, along with adding an adaptation stage, can substantially improve accuracy in few-shot settings, particularly for tasks within the Masked Filling category. As presented in Table \ref{table:results-kb}, this methodology shows a substantial improvement from the baseline for the Masked Filling category. \footnote{To ensure a fair comparison and to conduct experiments on separate categories, we reproduced the results on selected tasks using the settings specified in \cite{analyzingCS}. Our average score across all tasks closely matches the total average reported in the article.} Both the template and the denoising objective are important and contribute to achieving these results. 

The best performance is achieved by the MaskedPrompting and Mixed objectives during the adaptation phase. The denoising objective, which aligns with MaskedPrompting, ranks second. The Mixed approach, which adds Denoising and LM objectives based on the input, was discussed in Section \ref{sec:splitter}.

In the Map-Phrasal category, the advantage of the aligned adaptation phase, considering the nature of these tasks, is evident. Since the head and tail in this category already complement each other (e.g., PersonX cooks | to satisfy hunger), Mapping and Prompting act similarly. However, the LM or Mixed objective combined with Mapping outperforms others, followed by Denoising and MaskedPrompting, which are effective in both categories.

\begin{table}[htbp]
	\renewcommand{\arraystretch}{1.3}
	\centering
	\begin{tabular}{lcccccc}
		\toprule
		\textbf{Task Category} & \textbf{Method} & \textbf{Adaptation} & \textbf{Template} & \textbf{ROUGE} & \textbf{BERT Score} \\
		\midrule
		\multirow{4}{*}{Mask-Filling} & \multirow{2}{*}{Prompt Tuning \cite{lester2021power}} & -- & PrePT & 6.68 & 37.88 \\
		& & -- & PostPT & 5.68 & 35.6 \\ 
		\cline{2-6}
		& \multirow{3}{*}{MTO (ours)} & -- & MaskedPrePT & 7.40 & 34.89 \\ 
		& & -- & MaskedPostPT & 9.46 & 38.78 \\ 
		& & Denoising & MaskedPrePT & 14.93 & 45.19 \\
		& & \textbf{Denoising} & \textbf{MaskedPostPT} & \textbf{16.65} & \textbf{46.21} \\
		\midrule
		\multirow{4}{*}{Map-Phrasal}  
		& \multirow{2}{*}{Prompt Tuning}  & -- & PrePT & 37.26 & 48.14 \\
		& & -- & PostPT & 37.98 & 51.77 \\
		\cline{2-6}
		& \multirow{3}{*}{MTO (ours)} & -- & MaskedPrePT & 39.89 & 48.55  \\
		& & -- & MaskedPostPT & 39.93 & 51.59 \\
		& & LM & PrePT & 40.58 & 53.77 \\
		& & \textbf{Denoising} & \textbf{MaskedPrePT} & \textbf{41.98} & \textbf{53.80} \\
		\bottomrule
	\end{tabular}
	\caption{Comparison of our proposed methods' scores with the baseline (Prompt Tuning \cite{analyzingCS}) in few-shot settings, using $n=30$ samples on selected tasks from the $ATOMIC^{20}_{20}$ dataset.}
	\label{table:results-kb-pt}
\end{table}

\subsubsection{Prompt Tuning}
Table \ref{table:results-kb-pt} presents the results of KB completion tasks using methods proposed for prompt tuning, with our baseline being vanilla prompt-tuning \cite{lester2021power}. The positive impact of employing an adaptation phase aligned with the prompt-tuning method is evident. Similar to fine-tuning, we observe a trend where MaskedPostPT combined with the Denoising objective yields the best results for Mask Filling tasks. Importantly, the inclusion of soft, learnable prompts at the end of the input eliminates the need for hard prompt engineering, utilizing natural words. Moreover, with significantly reduced parameters, this method achieves performance close to that of the fine-tuned counterpart.

For Map-Phrasal tasks, the advantage of employing an adaptation phase and using a suitable template is evident once again. Approaches like PrePt combined with LM or PreMaskedPT, where prompts are positioned at the beginning and natural prompts are utilized before masked tokens at the end, demonstrate effective performance. MaskedPostPT also proves effective in this scenario too.

\begin{table*}[h!]
	\centering
	\begin{tabular}{lc|cc|c|cc}
		\toprule
		\textbf{Methods} & \multicolumn{3}{c}{\textbf{CommonSenseQA}} & \multicolumn{3}{c}{\textbf{OpenBookQA}} \\
		\cmidrule(lr){2-4} \cmidrule(lr){5-7}
		& \makecell{FS \\ 30} & \makecell{5\% \\ 425} & \makecell{20\% \\ 1700} & \makecell{FS \\ 30} & \makecell{5\% \\ 298} & \makecell{20\% \\ 991} \\
		\midrule
		RoBERTa-large \cite{roberta}    & -- & 29.66 & 58.47 & -- & 37.00 & 41.47 \\
		MHGRN \cite{base-MHGRN}          & -- & 29.01 & 50.23 & -- & 38.00 & 39.73 \\
		QA-GNN \cite{base-QAGNN}         & -- & 32.95 & 50.15 & -- & 33.53 & 42.40 \\
		GreaseLM \cite{base-GreaseLM}       & -- & 22.80 & 63.09 & -- & 39.00 & 42.20 \\
		GSC \cite{base-GSC}            & -- & 31.02 & 65.83 & -- & 29.60 & 42.40 \\
		SAFE  \cite{base-SAFE}           & -- & 36.45 & 65.16 & -- & 38.80 & 44.93 \\
		MVP-Tuning \cite{mvp}     & -- & 48.99 & 67.12 & -- & 39.60 & 56.00 \\					         
		\midrule
		\multicolumn{7}{c}{\textbf{Our Methods (Fine-Tuning)}} \\
		\midrule   
		T5-large (baseline)       & 23.94 & 23.12 & 65.15 & 23.73 & 49.21 & 55.31 \\      
		MTO-MP           & 20.09 & 25.53 & 59.02 & 23.93 & \textbf{53.60} & $\mathbf{57.52}^d$ \\  
		
		MTO-MCP          & $\textbf{52.5}_c^m$ &  $60.69^{d}$ & \textbf{68.12} & $\textbf{38.47}^l$ & 50.68 & $54.3^{l}$ \\ 
		
		MTO-AP           & $ 49.28^m$ & \textbf{$\mathbf{61.22}^d$} & $66.25^d$ & 32.73 & $52.2$ & $53.0^l$ \\  
		
		\midrule
		\multicolumn{7}{c}{\textbf{Our Methods (Prompt-Tuning)}} \\
		\midrule
		MTO-PreMAP          & $35.28^d$ & $ \underline{52.35}^d$ & $ \underline{59.21}^l$ & $25.23^l$ & $34.33^l$ & $ 39.65^l $ \\  
		MTO-PreAP          & $19.14^d$ & $33.41^l$ & $59.57$ & $4.2^d$ & $32.71^l$ & $32.42^l$ \\  
		\bottomrule
	\end{tabular}
	\caption{Comparison of the proposed methods with related works on the CommonSenseQA and OpenBookQA datasets. Superscripts denote the objectives used during the adaptation phase: \textit{d} for Denoising, \textit{l} for LM, and \textit{m} for Mixed. Additionally, MP in method names stands for Masked Prompting, MCP for Masked Choice Prompting, MAP for Masked Answer Prompting, and AP for Answer Prompting. In few-shot settings (FS $n=30$), the experiments were repeated three times, and the mean scores were reported. Detailed scores can be found in Appendix \ref{apx:results-fs}}
	\label{table:results-qa}
\end{table*}

\subsection{Question Answering}
Table \ref{table:results-qa} compares the performance of our methods with related works. In QA tasks, distinguishing the correct answer is essential, and this can be achieved using discriminative models. Therefore, the majority of the related works we examined employ encoder-only models, such as RoBERTa \cite{roberta}. These works typically utilize external knowledge from knowledge graphs to incorporate relevant information for each question to identify the correct choice. An overview of each of these works is presented in Appendix \ref{apx:baselines}. 

The works that use encoder-decoder models include \cite{unified} and \cite{rainbow}, which fine-tune T5 on various QA datasets or knowledge graphs in a multitasking fashion before applying it to a target dataset. However, their released models were already trained on CommonsenseQA, and cannot be tested in few-shot settings for direct comparison. Moreover, unlike these works, our methods primarily rely on the information within the model itself or unsupervised pretraining. 

As observed from the table, the performance of RoBERTa-large in low-resource settings (below 5\%) is higher than that of the T5 model. Using the full dataset, RoBERTa-large again outperforms T5 (refer to Table \ref{table:full}). However, we were able to substantially boost the performance of our base model beyond that of all the related works.

As observed, our proposed methods outperform the related works across all data regimes, particularly in few-shot settings. They also significantly surpass the baseline, which is the fine-tuned T5 model. This indicates that our methods effectively leverage the pre-trained knowledge of T5 to solve tasks without requiring extensive training data or additional information from knowledge graphs. This advantage is especially pronounced in the case of CommonsenseQA, which aligns well with the data and its structure within our adaptation stage. Nevertheless, our proposed methods perform well on both tasks in few-shot settings and consistently outperform the related works.

The advantage of predicting the choice number alongside the choice's text is evident in the results, with MaskedChoicePrompting (MTO-MCP) consistently outperforming other methods. This approach performs particularly well on CommonsenseQA, which features shorter answers, especially when the adaptation objective involves denoising.

With more training data (case 20\%), the gap between our scores and the baseline (T5) decreases. This demonstrates that training the model on a larger number of examples leads it to rely more on the format and knowledge within those examples. Additionally, as training samples increase, the impact of out-of-scope errors mentioned in Section \ref{sec:out-of-scope} can outweigh its initial advantage in detecting the correct choice. Furthermore, the model has limited capacity, and fine-tuning itself can lead to the forgetting of pre-trained knowledge and the mechanisms proposed to leverage them. Nevertheless, our proposed method, MTO-MP (MaskedPrompting), which places the question at the end of the input followed by the masked token, outperforms the baseline T5 in this case, which uses the typical format of placing the question at the start without employing masked tokens.  

\subsubsection{Results with the full dataset}
Table \ref{table:full} presents the performance results using the full dataset. It can be observed that our proposed method and the baseline achieve similar performance levels, although our method consistently outperforms the baseline, especially noticeable in the case of OpenBookQA. Therefore, not only are the proposed methods highly effective in low-resource settings, but they also demonstrate their usefulness in scenarios with ample resources.

\begin{table}[h!]
	\centering
	\begin{tabular}{lcc}
		\toprule
		& \textbf{CommonsenseQA} & \textbf{OpenBookQA} \\
		\midrule
		\textbf{Roberta-large (fine-tuned)} & 72.1 & 64.8 \\
		\textbf{T5-large (fine-tuned)} & 71.25$^\star$ & 61.8$^\star$ 	\\
		\midrule	
		\textbf{MTO-MP} & 68.96 & 64.8 \\
		\textbf{MTO-MAP} & \underline{71.74} & 62.8 \\
		\textbf{MTO-D+MCP} & 71.58 & 64.0 \\
		\textbf{MTO-AP} & 69.0 & \textbf{65.2} \\
		\bottomrule
	\end{tabular}
	\caption{Performance comparison of methods with the baseline using full data on CommonsenseQA and OpenBookQA. An asterisk (*) indicates fine-tuning performed using our settings. For reference, \citet{rel-dist} report a score of 62 for T5-large on OpenBookQA, and \citet{rel-calm} report a score of 69.81 for T5-large on CommonsenseQA. D+MCP stands for the Masked Choice Prompting method with a Denoising objective during the adaptation phase. MP: Masked Prompting, AP: Answer Prompting, and MAP: Masked Answer Prompting.}
	\label{table:full}
\end{table}

\subsubsection{Prompt-Tuning}
Table \ref{table:results-qa} also shows the prompt-tuning results. It is observed that PreMaskedAnswerPT (MTO-PreMAP), which uses masked tokens after the question and places the prompts at the beginning of sentences, can achieve high performance. Even in low-resource settings, prompt-tuning outperforms the related works and is on par with fine-tuning. Considering that in prompt-tuning a unique model can be applied to different tasks by plugging in prompts, and the advantage of faster and more efficient training, prompt-tuning can be an alternative in specific scenarios.

\section{Conclusion and Future Works}
In this study, we have explored techniques for enhancing language model performance through both full-model fine-tuning and prompt-tuning. Our findings emphasize the importance of aligning task formatting with pre-training objectives in both fine-tuning and prompt-tuning strategies. Once the objective that best matches the given task is identified, we adapt the model towards this objective by extending unsupervised training on a related corpus. If a model lacks a certain objective, it is beneficial to balance and complement the pre-training objectives to improve the model's efficiency across various tasks. 

We introduced a framework to prepare task-related data for adapting a model to specific tasks. This strategy involves identifying words or phrases that can be excluded from sentences within a corpus, enabling concurrent training on both denoising and Language Model (LM) objectives. Additionally, we introduced an automatic sentence splitter that learns to segment raw sentences based on examples from the target task. Furthermore, we proposed an objective classifier to determine the suitable objective for each sentence, thereby providing an instance-specific objective for the model. In the next stage, we proposed templates aligned with the objective used in the adaptation phase and demonstrated how this alignment can enhance performance.

In the current study, we utilized 8000 sentences during the adaptation phase. Exploring scenarios with larger training datasets remains a subject for future research.

These strategies lead to significant improvements in KB completion and QA tasks in few-shot settings and surpassing the baseline in the full dataset setting. This underscores how this alignment can effectively harness the embedded knowledge of a pre-trained language model (PLM).

The insights gleaned from this study offer valuable guidance for the selection and design of pre-training, fine-tuning, and prompt-tuning techniques across a spectrum of natural language processing tasks.

\subsection{Limitations}
Here are a few limitations within the current scope of this paper. First, while we discussed the advantages of the encoder-decoder architecture in generative tasks, our results are specific to this architecture. Future work could explore alternative architectures and pre-training objectives, and investigate how these models scale with model parameters and varying amounts of training data during the adaptation phase.

Secondly, our primary focus was on KB completion and question answering tasks. Exploring the impact of the proposed objective on classification and other task types warrants further investigation, which we intend to address in future studies.

\section{Declarations}
\begin{itemize}
	\item\textbf{Funding:} The authors did not receive support from any organization for the submitted work.
	\item  \textbf{Competing Interests:} This document is co-authored by Ahmad Pouramini and Hesham Faili. Ahmad Pouramini is a Ph.D. student at Tehran University and is employed as an instructor in the Computer Engineering Department at Sirjan University of Technology. Hesham Faili is a full-time Professor in the Electrical and Computer Engineering Department at Tehran University. We acknowledge the potential for conflicts of interest that may arise from our affiliations.
	
	\item \textbf{Code Availability:} We will publicly release the code used in this manuscript on GitHub at https://github.com/puraminy/MTO
	
\end{itemize}

\bibliography{sn-bibliography}% common bib file

\begin{thebibliography}{51}
\providecommand{\natexlab}[1]{#1}
\providecommand{\url}[1]{{#1}}
\providecommand{\urlprefix}{URL }
\providecommand{\doi}[1]{\url{https://doi.org/#1}}
\providecommand{\eprint}[2][]{\url{#2}}
 \bibcommenthead

\bibitem[{Bosselut et~al(2020{\natexlab{a}})Bosselut, Harrison, Anastasopoulos,
  Bisk, Hitzler, Kasture, and Riedel}]{comet2020}
Bosselut A, Harrison A, Anastasopoulos A, et~al (2020{\natexlab{a}})
  Comet-atomic 2020: On symbolic and neural commonsense knowledge graphs. In:
  Proceedings of the 34th AAAI Conference on Artificial Intelligence (AAAI
  2020), pp 6612--6619,
  \urlprefix\url{https://www.aaai.org/Papers/AAAI/2020GB/AAAI-BosselutA.6612.pdf}

\bibitem[{Bosselut et~al(2020{\natexlab{b}})Bosselut, Rashkin, Sap, Malaviya,
  Celikyilmaz, and Choi}]{comet}
Bosselut A, Rashkin H, Sap M, et~al (2020{\natexlab{b}}) {CoMET: Commonsense
  transformers for automatic knowledge graph construction}. In: ACL 2019 - 57th
  Annual Meeting of the Association for Computational Linguistics, Proceedings
  of the Conference. Association for Computational Linguistics, Florence,
  Italy, pp 4762--4779, \doi{10.18653/v1/p19-1470},
  \urlprefix\url{https://www.aclweb.org/anthology/P19-1470},
  \eprint{1906.05317}

\bibitem[{Brown et~al(2020)Brown, Mann, Ryder, Subbiah, Kaplan, Dhariwal,
  Neelakantan, Shyam, Sastry, Askell et~al}]{brown}
Brown T, Mann B, Ryder N, et~al (2020) Language models are few-shot learners.
  Advances in neural information processing systems 33:1877--1901

\bibitem[{Cao et~al(2021)Cao, Lin, Han, Sun, Yan, Liao, Xue, and Xu}]{rel-cao}
Cao B, Lin H, Han X, et~al (2021) Knowledgeable or educated guess? revisiting
  language models as knowledge bases. In: Proceedings of the 59th Annual
  Meeting of the Association for Computational Linguistics and the 11th
  International Joint Conference on Natural Language Processing (Volume 1: Long
  Papers). Association for Computational Linguistics, Online, pp 1860--1874,
  \doi{10.18653/v1/2021.acl-long.146},
  \urlprefix\url{https://aclanthology.org/2021.acl-long.146}

\bibitem[{Cao et~al(2024)Cao, Lin, Han, and Sun}]{k-survey}
Cao B, Lin H, Han X, et~al (2024) The life cycle of knowledge in big language
  models: A survey. Machine Intelligence Research pp 1--22

\bibitem[{Da et~al(2021)Da, Bras, Lu, Choi, and Bosselut}]{analyzingCS}
Da J, Bras RL, Lu X, et~al (2021) Analyzing commonsense emergence in few-shot
  knowledge models. In: Conference on Automated Knowledge Base Construction,
  \urlprefix\url{https://api.semanticscholar.org/CorpusID:235657379}

\bibitem[{Feldman et~al(2020)Feldman, Davison, and Rush}]{davison-cs-mining}
Feldman J, Davison J, Rush AM (2020) {Commonsense knowledge mining from
  pretrained models}. In: EMNLP-IJCNLP 2019 - 2019 Conference on Empirical
  Methods in Natural Language Processing and 9th International Joint Conference
  on Natural Language Processing, Proceedings of the Conference. Association
  for Computational Linguistics, Hong Kong, China, pp 1173--1178,
  \doi{10.18653/v1/d19-1109},
  \urlprefix\url{https://www.aclweb.org/anthology/D19-1109},
  \eprint{1909.00505}

\bibitem[{Feng et~al(2020)Feng, Chen, Lin, Wang, Yan, and Ren}]{base-MHGRN}
Feng Y, Chen X, Lin BY, et~al (2020) Scalable multi-hop relational reasoning
  for knowledge-aware question answering. In: Proceedings of the 2020
  Conference on Empirical Methods in Natural Language Processing (EMNLP),
  Association for Computational Linguistics, Online, pp 1295--1309

\bibitem[{Fichtel et~al(2021)Fichtel, Kalo, and Balke}]{rel-fin1}
Fichtel L, Kalo JC, Balke WT (2021) Prompt tuning or fine-tuning -
  investigating relational knowledge in pre-trained language models. In: 3rd
  Conference on Automated Knowledge Base Construction, \doi{10.24432/C5RC75},
  \urlprefix\url{https://openreview.net/forum?id=o7sMlpr9yBW}

\bibitem[{Gururangan et~al(2020)Gururangan, Marasovi{\'c}, Swayamdipta, Lo,
  Beltagy, Downey, and Smith}]{dontstop}
Gururangan S, Marasovi{\'c} A, Swayamdipta S, et~al (2020) Don't stop
  pretraining: Adapt language models to domains and tasks. In: Proceedings of
  the 58th Annual Meeting of the Association for Computational Linguistics,
  Association for Computational Linguistics, Online, pp 8342--8360

\bibitem[{Hase et~al(2023)Hase, Diab, Celikyilmaz, Li, Kozareva, Stoyanov,
  Bansal, and Iyer}]{rel-cs}
Hase P, Diab M, Celikyilmaz A, et~al (2023) Methods for measuring, updating,
  and visualizing factual beliefs in language models. In: Vlachos A, Augenstein
  I (eds) Proceedings of the 17th Conference of the European Chapter of the
  Association for Computational Linguistics. Association for Computational
  Linguistics, Dubrovnik, Croatia, pp 2714--2731,
  \doi{10.18653/v1/2023.eacl-main.199},
  \urlprefix\url{https://aclanthology.org/2023.eacl-main.199}

\bibitem[{He et~al(2024)He, Fang, Wang, and Song}]{concept}
He M, Fang T, Wang W, et~al (2024) Acquiring and modeling abstract commonsense
  knowledge via conceptualization. Artificial Intelligence p 104149

\bibitem[{Huang et~al(2023)Huang, Li, Xu, Zhang, Gan, Zhang, and Wang}]{mvp}
Huang Y, Li Y, Xu Y, et~al (2023) Mvp-tuning: Multi-view knowledge retrieval
  with prompt tuning for commonsense reasoning. In: Proceedings of the 61st
  Annual Meeting of the Association for Computational Linguistics (Volume 1:
  Long Papers), pp 13417--13432

\bibitem[{Jiang et~al(2020)Jiang, Xu, Araki, and Neubig}]{rel-jiang}
Jiang Z, Xu FF, Araki J, et~al (2020) How can we know what language models
  know? Transactions of the Association for Computational Linguistics
  8:423--438.
  \urlprefix\url{https://www.mitpressjournals.org/doi/abs/10.1162/tacl_a_00323}

\bibitem[{Kang et~al(2023)Kang, Lee, Baek, Kawaguchi, and Hwang}]{rel-dist}
Kang M, Lee S, Baek J, et~al (2023) Knowledge-augmented reasoning distillation
  for small language models in knowledge-intensive tasks. In: Oh A, Naumann T,
  Globerson A, et~al (eds) Advances in Neural Information Processing Systems,
  vol~36. Curran Associates, Inc., pp 48573--48602,
  \urlprefix\url{https://proceedings.neurips.cc/paper_files/paper/2023/file/97faedc90260eae5c400f92d5831c3d7-Paper-Conference.pdf}

\bibitem[{Kazemi et~al(2023)Kazemi, Mittal, and Ramachandran}]{rel-fin2}
Kazemi M, Mittal S, Ramachandran D (2023) Understanding finetuning for factual
  knowledge extraction from language models. arXiv e-prints pp arXiv--2301

\bibitem[{Khashabi et~al(2020)Khashabi, Min, Khot, Sabharwal, Tafjord, Clark,
  and Hajishirzi}]{unified}
Khashabi D, Min S, Khot T, et~al (2020) Unifiedqa: Crossing format boundaries
  with a single qa system. In: Findings of the Association for Computational
  Linguistics: EMNLP 2020. Association for Computational Linguistics, pp
  1896--1907, \urlprefix\url{https://aclanthology.org/2020.findings-emnlp.171}

\bibitem[{Lester et~al(2021)Lester, Al-Rfou, and Constant}]{lester2021power}
Lester B, Al-Rfou R, Constant N (2021) The power of scale for
  parameter-efficient prompt tuning. In: Proceedings of the 2021 Conference on
  Empirical Methods in Natural Language Processing. Association for
  Computational Linguistics, Online and Punta Cana, Dominican Republic, pp
  3045--3059, \doi{10.18653/v1/2021.emnlp-main.243},
  \urlprefix\url{https://aclanthology.org/2021.emnlp-main.243}

\bibitem[{Li et~al(2023)Li, Wang, Chen, Liu, and Zhao}]{springer-event}
Li J, Wang C, Chen Y, et~al (2023) What events do pre-trained language models
  learn from text? probing event-based commonsense knowledge by confidence
  sorting. In: Liu F, Duan N, Xu Q, et~al (eds) Natural Language Processing and
  Chinese Computing. Springer Nature Switzerland, Cham, pp 669--681

\bibitem[{Li and Liang(2021)}]{prefix}
Li XL, Liang P (2021) {Prefix-tuning: Optimizing continuous prompts for
  generation}. ACL-IJCNLP 2021 - 59th Annual Meeting of the Association for
  Computational Linguistics and the 11th International Joint Conference on
  Natural Language Processing, Proceedings of the Conference pp 4582--4597.
  \doi{10.18653/v1/2021.acl-long.353},
  {\href{https://arxiv.org/abs/2101.00190}{{2101.00190}}}

\bibitem[{Lin(2004)}]{rouge}
Lin CY (2004) Rouge: A package for automatic evaluation of summaries. In:
  Annual Meeting of the Association for Computational Linguistics,
  \urlprefix\url{https://api.semanticscholar.org/CorpusID:964287}

\bibitem[{Liu and Singh(2004)}]{omcs}
Liu H, Singh P (2004) Conceptnet: A practical commonsense reasoning toolkit.
  In: BT technology journal, vol~22. Springer, pp 211--226

\bibitem[{Liu et~al(2023)Liu, Yuan, Fu, Jiang, Hayashi, and Neubig}]{ppp}
Liu P, Yuan W, Fu J, et~al (2023) Pre-train, prompt, and predict: A systematic
  survey of prompting methods in natural language processing. ACM Computing
  Surveys 55(9):1--35

\bibitem[{Liu et~al(2021)Liu, Zheng, Du, Ding, Qian, Yang, and Tang}]{gptund}
Liu X, Zheng Y, Du Z, et~al (2021) Gpt understands, too. CoRR abs/2103.10385

\bibitem[{Liu et~al(2019)Liu, Ott, Goyal, Du, Joshi, Chen, Levy, Lewis,
  Zettlemoyer, and Stoyanov}]{roberta}
Liu Y, Ott M, Goyal N, et~al (2019) Roberta: A robustly optimized bert
  pretraining approach. arXiv preprint arXiv:190711692

\bibitem[{Lourie et~al(2021)Lourie, Le~Bras, Bhagavatula, and Choi}]{rainbow}
Lourie N, Le~Bras R, Bhagavatula C, et~al (2021) Unicorn on rainbow: A
  universal commonsense reasoning model on a new multitask benchmark. In:
  Proceedings of the AAAI Conference on Artificial Intelligence, pp
  13480--13488

\bibitem[{Mihaylov et~al(2018)Mihaylov, Clark, Khot, and Sabharwal}]{openbook}
Mihaylov T, Clark P, Khot T, et~al (2018) Can a suit of armor conduct
  electricity? a new dataset for open book question answering. In: Proceedings
  of the 2018 Conference on Empirical Methods in Natural Language Processing.
  Association for Computational Linguistics, Brussels, Belgium, pp 2381--2391,
  \doi{10.18653/v1/D18-1260}, \urlprefix\url{https://aclanthology.org/D18-1260}

\bibitem[{Petroni et~al(2020)Petroni, Rockt{\"{a}}schel, Lewis, Bakhtin, Wu,
  Miller, and Riedel}]{lama}
Petroni F, Rockt{\"{a}}schel T, Lewis P, et~al (2020) {Language models as
  knowledge bases?} In: EMNLP-IJCNLP 2019 - 2019 Conference on Empirical
  Methods in Natural Language Processing and 9th International Joint Conference
  on Natural Language Processing, Proceedings of the Conference. Association
  for Computational Linguistics, Hong Kong, China, pp 2463--2473,
  \doi{10.18653/v1/d19-1250},
  \urlprefix\url{https://www.aclweb.org/anthology/D19-1250},
  \eprint{1909.01066}

\bibitem[{Qin and Eisner(2021)}]{rel-contprompt2}
Qin G, Eisner J (2021) Learning how to ask: Querying lms with mixtures of soft
  prompts. In: Proceedings of the 2021 Conference of the North American Chapter
  of the Association for Computational Linguistics: Human Language
  Technologies, pp 5203--5212

\bibitem[{Raffel et~al(2020)Raffel, Shazeer, Roberts, Lee, Narang, Matena,
  Zhou, Li, and Liu}]{t5}
Raffel C, Shazeer N, Roberts A, et~al (2020) Exploring the limits of transfer
  learning with a unified text-to-text transformer. Journal of Machine Learning
  Research 21(140):1--67.
  \urlprefix\url{http://jmlr.org/papers/v21/20-074.html}

\bibitem[{Sanh et~al(2022)Sanh, Webson, Raffel, Bach, Sutawika, Alyafeai,
  Chaffin, Stiegler, Le~Scao, Raja et~al}]{t0}
Sanh V, Webson A, Raffel C, et~al (2022) Multitask prompted training enables
  zero-shot task generalization. In: ICLR 2022-Tenth International Conference
  on Learning Representations

\bibitem[{Sap et~al(2019)Sap, {Le Bras}, Allaway, Bhagavatula, Lourie, Rashkin,
  Roof, Smith, and Choi}]{atomic}
Sap M, {Le Bras} R, Allaway E, et~al (2019) {ATOMIC: An atlas of machine
  commonsense for if-then reasoning}. In: 33rd AAAI Conference on Artificial
  Intelligence, AAAI 2019, 31st Innovative Applications of Artificial
  Intelligence Conference, IAAI 2019 and the 9th AAAI Symposium on Educational
  Advances in Artificial Intelligence, EAAI 2019, pp 3027--3035,
  \doi{10.1609/aaai.v33i01.33013027}, \eprint{1811.00146}

\bibitem[{Schick and Sch{\"{u}}tze(2021)}]{rel-timo}
Schick T, Sch{\"{u}}tze H (2021) {Few-Shot Text Generation with Natural
  Language Instructions}. EMNLP 2021 - 2021 Conference on Empirical Methods in
  Natural Language Processing, Proceedings pp 390--402.
  \doi{10.18653/v1/2021.emnlp-main.32}

\bibitem[{Shazeer and Stern(2018)}]{ada}
Shazeer N, Stern M (2018) Adafactor: Adaptive learning rates with sublinear
  memory cost. In: International Conference on Machine Learning, PMLR, pp
  4596--4604

\bibitem[{Shin et~al(2020)Shin, Razeghi, Logan~IV, Wallace, and
  Singh}]{rel-shin}
Shin T, Razeghi Y, Logan~IV RL, et~al (2020) Autoprompt: Eliciting knowledge
  from language models with automatically generated prompts. In: Proceedings of
  the 2020 Conference on Empirical Methods in Natural Language Processing
  (EMNLP), Association for Computational Linguistics, pp 4222--4235,
  \urlprefix\url{https://www.aclweb.org/anthology/2020.emnlp-main.343/}

\bibitem[{Talmor et~al(2019)Talmor, Herzig, Lourie, and Berant}]{commonsenseqa}
Talmor A, Herzig J, Lourie N, et~al (2019) {C}ommonsense{QA}: A question
  answering challenge targeting commonsense knowledge. In: Proceedings of the
  2019 Conference of the North American Chapter of the Association for
  Computational Linguistics: Human Language Technologies, Volume 1 (Long and
  Short Papers). Association for Computational Linguistics, Minneapolis,
  Minnesota, pp 4149--4158, \doi{10.18653/v1/N19-1421},
  \urlprefix\url{https://aclanthology.org/N19-1421}

\bibitem[{Wallat et~al(2020)Wallat, Singh, and Anand}]{rel-wallet}
Wallat J, Singh J, Anand A (2020) Bertnesia: Investigating the capture and
  forgetting of knowledge in bert. In: Proceedings of the Third BlackboxNLP
  Workshop on Analyzing and Interpreting Neural Networks for NLP, pp 174--183

\bibitem[{Wang et~al(2021)Wang, Zhang, Yang, Song, and Qin}]{base-GSC}
Wang K, Zhang Y, Yang D, et~al (2021) Gnn is a counter? revisiting gnn for
  question answering. In: International Conference on Learning Representations

\bibitem[{Wang et~al(2022)Wang, Roberts, Hesslow, Le~Scao, Chung, Beltagy,
  Launay, and Raffel}]{related-what}
Wang T, Roberts A, Hesslow D, et~al (2022) What language model architecture and
  pretraining objective works best for zero-shot generalization? In:
  International Conference on Machine Learning, PMLR, pp 22964--22984

\bibitem[{Wangchunshu~Zhou(2021)}]{rel-calm}
Wangchunshu~Zhou RKSSLBYLXRDong-Ho~Lee (2021) Pre-training text-to-text
  transformers for concept-centric common sense. In: International Conference
  on Learning Representations (ICLR)

\bibitem[{West et~al(2022)West, Bhagavatula, Hessel, Hwang, Jiang, Le~Bras, Lu,
  Welleck, and Choi}]{west}
West P, Bhagavatula C, Hessel J, et~al (2022) Symbolic knowledge distillation:
  from general language models to commonsense models. In: Carpuat M,
  de~Marneffe MC, Meza~Ruiz IV (eds) Proceedings of the 2022 Conference of the
  North American Chapter of the Association for Computational Linguistics:
  Human Language Technologies. Association for Computational Linguistics,
  Seattle, United States, pp 4602--4625, \doi{10.18653/v1/2022.naacl-main.341},
  \urlprefix\url{https://aclanthology.org/2022.naacl-main.341}

\bibitem[{Yasunaga et~al(2021)Yasunaga, Ren, Bosselut, Liang, and
  Leskovec}]{base-QAGNN}
Yasunaga M, Ren H, Bosselut A, et~al (2021) Qa-gnn: reasoning with language
  models and knowledge graphs for question answering. In: Proceedings of the
  2021 Conference of the North American Chapter of the Association for
  Computational Linguistics: Human Language Technologies, Association for
  Computational Linguistics, Online, pp 535--546

\bibitem[{Yin et~al(2022)Yin, Bansal, Monajatipoor, Li, and Chang}]{rel-yin}
Yin D, Bansal H, Monajatipoor M, et~al (2022) Geomlama: Geo-diverse commonsense
  probing on multilingual pre-trained language models. In: Proceedings of the
  2022 Conference on Empirical Methods in Natural Language Processing, pp
  2039--2055

\bibitem[{Zhang et~al(2022{\natexlab{a}})Zhang, Liu, Pan, Ke, Ou, Fang, and
  Song}]{aser}
Zhang H, Liu X, Pan H, et~al (2022{\natexlab{a}}) Aser: Towards large-scale
  commonsense knowledge acquisition via higher-order selectional preference
  over eventualities. Artificial Intelligence 309:103740

\bibitem[{Zhang and Li(2024)}]{springer-cs}
Zhang L, Li R (2024) Knowledge prompting with contrastive learning
  for unsupervised commonsenseqa. In: Luo B, Cheng L, Wu ZG, et~al (eds)
  Neural Information Processing. Springer Nature Singapore, Singapore, pp
  27--38

\bibitem[{Zhang et~al(2020)Zhang, Kishore, Wu, Weinberger, and
  Artzi}]{bertscore}
Zhang T, Kishore V, Wu F, et~al (2020) Bertscore: Evaluating text generation
  with bert. In: International Conference on Learning Representations (ICLR),
  \urlprefix\url{https://openreview.net/forum?id=SkeHuCVFDr}

\bibitem[{Zhang et~al(2022{\natexlab{b}})Zhang, Bosselut, Yasunaga, Ren, Liang,
  Manning, and Leskovec}]{base-GreaseLM}
Zhang X, Bosselut A, Yasunaga M, et~al (2022{\natexlab{b}}) Greaselm: Graph
  reasoning enhanced language models for question answering. In: International
  Conference on Representation Learning (ICLR)

\bibitem[{Zhao et~al(2022)Zhao, Jiang, Zhou, and Wen}]{base-SAFE}
Zhao WX, Jiang J, Zhou K, et~al (2022) Great truths are always simple: A rather
  simple knowledge encoder for enhancing the commonsense reasoning capacity of
  pre-trained models. In: Findings of the North American Chapter of the
  Association for Computational Linguistics: NAACL 2022, Association for
  Computational Linguistics

\bibitem[{Zhao et~al(2021)Zhao, Wallace, Feng, Klein, and Singh}]{rel-zhao}
Zhao Z, Wallace E, Feng S, et~al (2021) Calibrate before use: Improving
  few-shot performance of language models. In: International Conference on
  Machine Learning, PMLR, pp 12697--12706

\bibitem[{Zhong et~al(2021)Zhong, Friedman, and Chen}]{rel-contprompt}
Zhong Z, Friedman D, Chen D (2021) Factual probing is [mask]: Learning vs.
  learning to recall. In: Proceedings of the 2021 Conference of the North
  American Chapter of the Association for Computational Linguistics: Human
  Language Technologies, pp 5017--5033

\bibitem[{Zhou et~al(2020)Zhou, Zhang, Cui, and Huang}]{rel-zhou}
Zhou X, Zhang Y, Cui L, et~al (2020) Evaluating commonsense in pre-trained
  language models. In: Proceedings of the AAAI Conference on Artificial
  Intelligence, pp 9733--9740

\end{thebibliography}
%% if required, the content of .bbl file can be included here once bbl is generated
%%\input sn-article.bbl

\begin{appendices}
\section{Templates and Tasks}	
	\subsection{Templates for the relations}
	Table \ref{table:tuples} presents a compilation of natural language phrases that have been employed for formatting relation tuples into coherent natural language sentences. These phrases are used as part of the fine-tuning process to shape the inputs and outputs of the model for training.
	
	\begin{table*}[htbp]
		\centering
		\begin{tabular}{|p{2.5cm}|p{4cm}|p{6cm}|}
			\toprule
			\textbf{Relation} & \textbf{Natural phrase} & \textbf{Example} \\
			\midrule
			AtLocation & located at & Book is located at the library \\
			ObjectUse & is used for & Hammer is used for building \\
			CapableOf & is capable of & Athlete is capable of running \\
			HasProperty & has the property of & The car has the property of being fast \\
			isFilledBy & is filled by & PersonX watches ---  anyway is filled by the TV \\
			xAttr & is seen as & PersonX teaches at a university. PersonX is seen as intelligent \\     
			\midrule
			xIntent & because they intended & PersonX eats vegetables because they intended to be healthy \\
			xNeed & before that they need & PersonX attend the marathon, before that they need to train \\
			\midrule   
			QA & the correct choice is & Where can you find a restaurant's phone number? A) Yellow pages B) bath \\       
			\bottomrule
		\end{tabular}
		\caption{Natural language phrases used for formatting relations in natural language sentences}
		\label{table:tuples}
	\end{table*}
	
	\subsection{Task Categories for Training Objective Classifier}
	The categories of the selected tasks from the ATOMIC2020 dataset to train the objective classifier discussed in Section \ref{sec:classifier}
	\begin{table*}[h]
		\centering
		\begin{tabular}{ll}
			\toprule
			\textbf{Mask-Filling} & \textbf{Phrasal-Map} \\
			\midrule
			Desires & xIntent \\
			CapableOf & xWant \\
			xReact & xEffect \\
			xAttr & xNeed \\
			Causes & isAfter \\
			AtLocation & isBefore \\
			HasProperty & oWant \\
			ObjectUse & HasSubEvent \\
			MadeUpOf & \\
			\bottomrule
		\end{tabular}
		\caption{Categories of the selected tasks from ATOMIC2020 dataset to train the objective classifier \ref{sec:classifier}}
		\label{table:filling_mapping}
	\end{table*}

\section{Details of scores}	
	\subsection{Corelation between the scores of different metrics}\label{sec:appendix-bert}
	
\begin{figure}[h!]
		\centering
		\includegraphics[width=0.3\linewidth]{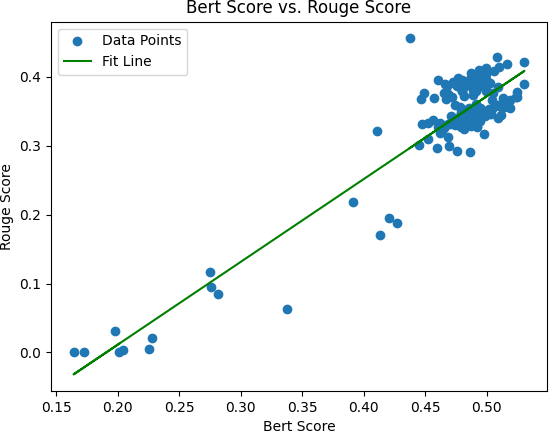}
		\includegraphics[width=0.3\linewidth]{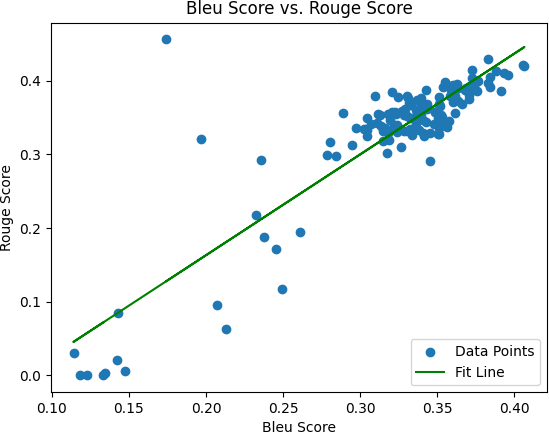}
		\includegraphics[width=0.3\linewidth]{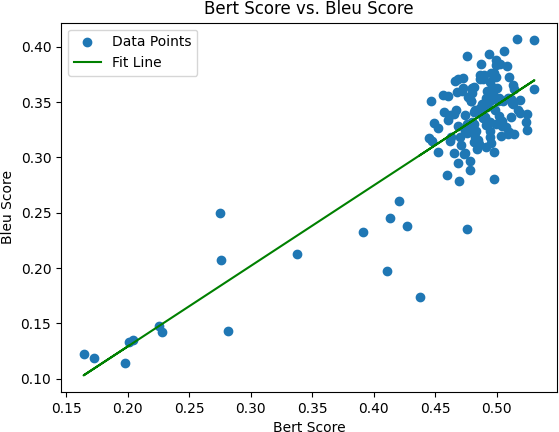}
		\caption{Correlation between different metrics (ROUGE, BERT, and BLEURT scores) for the Map-Phrasal tasks. Similar correlations were observed in other experiments as well.}
		\label{fig:rg-bert-bleu}
	\end{figure}

In our experiments, we observed a correlation between the BLEURT, ROUGE, and BERT scores. Comparing each prediction with three references contributes to the robustness of these scores. This correlation is evident in Figure \ref{fig:rg-bert-bleu} and Table \ref{table:top-ft-filling} and Table \ref{table:top-ft-map}.

	\subsection{Top Scores of Different models} \label{sec:scores}
	Table \ref{table:top-ft-filling} and Table \ref{table:top-ft-map} display the highest scores achieved by various models across three metrics: Bert Score, ROUGE, and BLEURT in few-shot settings.
	
	\begin{table*}[htbp]
		\renewcommand{\arraystretch}{1}
		\centering
		\begin{tabular}{lll|llllllll}
	\hline
	\multirow{2}{*}{model} & \multirow{2}{*}{objective} & \multirow{2}{*}{\rot{template}} & \multicolumn{7}{c}{Tasks} \\
	\cline{4-11}
	&                             &                                  & \rot{AtLocation} & \rot{CapableOf} & \rot{HasProperty} & \rot{ObjectUse} & \rot{isFilledBy} & \rot{xAttr} & Avg. \\
	\toprule
	\multirow{2}{*}{T5-large} & \multirow{2}{*}{Mixed}      & \multirow{2}{*}{MP}     & 0.54 & 0.50 & 0.40 & 0.53 & 0.46 & 0.55 & \textbf{0.50} & BERT    \\
	&                             &                               & 0.22 & 0.24 & 0.14 & 0.28 & 0.17 & 0.15 & \textbf{0.20} & ROUGE  \\
	\midrule
	\multirow{2}{*}{T5-base}  & \multirow{2}{*}{LM}         & \multirow{2}{*}{MP}         & 0.50 & 0.46 & 0.43 & 0.47 & 0.41 & 0.53 & 0.47 & BERT \\
	&                             &                                & 0.20 & 0.20 & 0.18 & 0.20 & 0.12 & 0.18 & 0.18 & ROUGE \\
	\midrule
	\multirow{2}{*}{T5-lm}    & \multirow{2}{*}{Denoising}  & \multirow{2}{*}{MP}         & 0.50 & 0.45 & 0.41 & 0.47 & 0.47 & 0.53 & 0.47 & BERT \\
	&                             &                                & 0.16 & 0.19 & 0.13 & 0.21 & 0.21 & 0.18 & 0.18 & ROUGE \\
	\midrule
	\multirow{2}{*}{T5-v1}    & \multirow{2}{*}{Denoising}  & \multirow{2}{*}{MP}         & 0.48 & 0.46 & 0.44 & 0.50 & 0.46 & 0.52 & 0.47 & BERT \\
	&                             &                                & 0.16 & 0.19 & 0.17 & 0.22 & 0.19 & 0.14 & 0.18 & ROUGE \\
	\bottomrule
\end{tabular}
 
		\caption{Top Rouge and BERT scores evaluating the tail generation quality across various models with distinct unsupervised objectives and fine-tuning templates, with $n = 30$ samples for each task in the Masked-Filling category, based on 3 epochs. MP stands for MaskedPrompting.}
		\label{table:top-ft-filling}
	\end{table*}

	\begin{table*}[th!]
	\renewcommand{\arraystretch}{1}
	\centering
	\begin{tabular}{llllllll}
\hline
 model   & objective   & template      & xIntent       & xNeed         & xWant         & All           & metric                  \\
\hline
\hline
T5-large & LM          & Mapping    & \textbf{0.57} & \textbf{0.57} & \textbf{0.55} & \textbf{0.56} & \multirow{1}{*}{BERT}   \\
T5-large & LM          & Mapping    & \textbf{0.48} & \textbf{0.43} & \textbf{0.38} & \textbf{0.43} & \multirow{1}{*}{Rouge}  \\
\hline
 T5-v1   & LM          & MaskedMapping & \textbf{0.51} & \textbf{0.50} & 0.50          & \textbf{0.50} & \multirow{4}{*}{BERT}   \\
 T5-base & Denoising   & Mapping       & 0.48          & \textbf{0.50} & \textbf{0.51} & \textbf{0.50} &                         \\
 T5-lm   & Denoising   & Mapping       & 0.50          & \textbf{0.50} & 0.49          & \textbf{0.50} &                         \\
 T5-base & ---         & Prompting     & 0.48          & 0.48          & 0.50          & 0.49          &                         \\
\hline
 T5-v1   & LM          & MaskedMapping & \textbf{0.38} & \textbf{0.33} & 0.28          & 0.33          & \multirow{4}{*}{BLEURT} \\
 T5-base & Denoising   & Mapping       & 0.36          & 0.30          & \textbf{0.37} & \textbf{0.34} &                         \\
 T5-lm   & Denoising   & Mapping       & 0.36          & \textbf{0.33} & 0.33          & \textbf{0.34} &                         \\
 T5-base & ---         & Prompting     & 0.35          & 0.31          & 0.35          & \textbf{0.34} &                         \\
\hline
 T5-v1   & LM          & MaskedMapping & \textbf{0.43} & \textbf{0.38} & 0.32          & \textbf{0.38} & \multirow{4}{*}{Rouge}  \\
 T5-base & Denoising   & Mapping       & 0.38          & 0.35          & \textbf{0.37} & 0.37          &                         \\
 T5-lm   & Denoising   & Mapping       & 0.39          & 0.37          & 0.35          & 0.37          &                         \\
 T5-base & ---         & Prompting     & 0.40          & 0.34          & 0.35          & 0.36          &                         \\
\hline
\end{tabular} 
	\caption{Top BERT, BLEURT and Rouge scores evaluating the tail generation quality across various models with distinct unsupervised objectives and fine-tuning templates, with $n = 30$ samples for each task in the Map-Phrasal category. MP stands for MaskedPrompting.}
	\label{table:top-ft-map}
    \end{table*}

	 \subsection{Fewshot Performance of QA tasks} \label{apx:results-fs}
Table \ref{table:results-fs} presents the mean and standard deviation of accuracy for QA datasets in few-shot settings. As observed, MaskedChoicePrompting consistently outperforms the other methods in both CommonSenseQA and OpenBookQA when the model undergoes the adaptation phase with the Denoising objective.

	\begin{table}[ht]
		\centering
	
		\begin{tabular}{llcc|cc}
			\toprule
			\multirow{2}{*}{\textbf{Adaptation}} & \multirow{2}{*}{\textbf{Template}} & \multicolumn{2}{c}{\textbf{CommonsenseQA}} & \multicolumn{2}{c}{\textbf{OpenBookQA}} \\
			& & \textbf{Mean} & \textbf{Std} & \textbf{Mean} & \textbf{Std} \\
			\midrule
			\multirow{6}{*}{Denoising} 
			& AnswerPrompting & 45.15 & $\pm$ 6.24 & 28.7 & $\pm$ 1.80 \\
			& ChoicePrompting & 27.03 & $\pm$ 4.12 & 29.07 & $\pm$ 5.19 \\
			& Mapping & 22.69 & $\pm$ 4.24 & 25.93 & $\pm$ 1.45 \\
			& MaskedAnswerPrompting & 44.55 & $\pm$ 2.79 & 23.87 & $\pm$ 0.76 \\
			& \textbf{MaskedChoicePrompting} & \textbf{50.86} & $\pm$ 2.17 & \textbf{33.85} & $\pm$ 3.20 \\
			& MaskedPrompting & 22.82 & $\pm$ 1.75 & 28.07 & $\pm$ 4.40 \\
			\midrule
			\multirow{6}{*}{LM} 
			& AnswerPrompting & 47.75 & $\pm$ 3.06 & 29.47 & $\pm$ 4.05 \\
			& ChoicePrompting & 46.49 & $\pm$ 1.87 & 32.6 & $\pm$ 2.88 \\
			& Mapping & 22.14 & $\pm$ 3.38 & 27.8 & $\pm$ 5.97 \\
			& MaskedAnswerPrompting & 45.21 & $\pm$ 2.99 & 27.8 & $\pm$ 2.03 \\
			& \textbf{MaskedChoicePrompting} & \textbf{52.55} & $\pm$ 4.00 & \textbf{38.47} & $\pm$ 0.95 \\
			& MaskedPrompting & 21.65 & $\pm$ 1.89 & 26.47 & $\pm$ 1.72 \\
			\midrule
			\multirow{6}{*}{Mixed} 
			& \underline{AnswerPrompting} &  \underline{49.28} & $\pm$ \underline{0.54} & 32.4 & $\pm$ 1.39 \\
			& ChoicePrompting & 38.27 & $\pm$ 9.17 & 28.67 & $\pm$ 4.50 \\
			& Mapping & 21.48 & $\pm$ 1.85 & 26.93 & $\pm$ 5.33 \\
			& MaskedAnswerPrompting & 46.38 & $\pm$ 2.42 & 24.13 & $\pm$ 2.91 \\
			& \textbf{MaskedChoicePrompting} & \textbf{52.5} & $\pm$ 2.20 & \textbf{34.53} & $\pm$ 2.48 \\
			& MaskedPrompting & 19.98 & $\pm$ 0.58 & 23.53 & $\pm$ 3.56 \\
			\midrule
			\multirow{5}{*}{--} 
			& \textbf{AnswerPrompting} & \textbf{49.74} & $\pm$ 3.93 & \textbf{32.73} & $\pm$ 1.97 \\
			& Mapping & 23.94 & $\pm$ 5.74 & 23.73 & $\pm$ 4.59 \\
			& MaskedAnswerPrompting & 45.59 & $\pm$ 4.45 & 25.93 & $\pm$ 3.36 \\
			& MaskedChoicePrompting & 46.44 & $\pm$ 6.19 & 31.6 & $\pm$ 4.28 \\
			& MaskedPrompting & 20.09 & $\pm$ 0.45 & 23.93 & $\pm$ 2.64 \\
			\bottomrule
		\end{tabular}
		\caption{Mean and standard deviation of accuracy across three runs with varying random seeds for CommonsenseQA and OpenBookQA, based on 3 epochs and $n=30 $ samples} 
		\label{table:results-fs}
	\end{table}

\section{Datasets}	
\subsection{QA Dataset Examples} \label{apx:ds-examples}
	Examples of the QA datasets CommonsenseQA and OpenBookQA include questions like Table \ref{table:commonsense} and Table \ref{table:openbook}.
		\begin{table}[h]
		\centering
		\begin{tabular}{p{10cm}}
			\toprule

			\textbf{Question:} What would you need if you want to smoke? \\
			A) you're stupid \\
			B) kill yourself \\
			C) roll joint \\
			D) cigarette \\
			E) lighter fluid \\ \midrule
			
			\textbf{Question:} Where can you find a restaurant's phone number? \\
			A) yellow pages \\
			B) city \\
			C) business sector \\
			D) town \\
			E) at hotel \\ \midrule
			
			\textbf{Question:} What uses a ribbon to put words on paper? \\
			A) wrapping paper \\
			B) girl's hair \\
			C) bath \\
			D) floral arrangement \\
			E) typewriter \\ \midrule
			
			\textbf{Question:} Lawyers often talk in front of an audience where? \\
			A) work \\
			B) courtroom \\
			C) office building \\
			D) press charges \\
			E) theatre \\ \midrule
			
			\textbf{Question:} James loved to surf but he wasn't good at it. He would always do what? \\
			A) wipe out \\
			B) enjoy yourself \\
			C) start fighting \\
			D) get wet \\
			E) drown \\ \bottomrule
			
		\end{tabular}
		\caption{Examples from the Commonsense QA Dataset}
		\label{table:commonsense}
	\end{table}

	\begin{table}[h!]
	\centering
	\begin{tabular}{p{10cm}}
		\midrule
		\textbf{Question:} The organelle that provides energy to the cell is the: \\
		A) Nucleus \\
		B) Mitochondria \\
		C) Ribosome \\
		D) Golgi apparatus \\ \midrule
		
		\textbf{Question:} Which of the following best explains why leaves are typically green? \\
		A) They absorb green light. \\
		B) They reflect green light. \\
		C) They convert green light to energy. \\
		D) They store green pigments. \\ \midrule
		
		\textbf{Question:}  shark will be unable to survive on eating algae and moss, because \\
		A) it is a predator \\
		B) it is a vegetarian  \\
		C) it is a freshwater fish \\
		D) it is a producer \\ \midrule
		
		\textbf{Question:} If bacon is left too long on a hot stove top... \\ 
		A) it will be cooked perfectly \\ 
		B) it will be bacteria laden \\ 
		C) it will become blackened \\
		D) it will be left raw \\ \midrule
		
	\end{tabular}
	\caption{Examples from the OpenBookQA Dataset}
	\label{table:openbook}
\end{table}

	\subsection{Analysis of Model Prediction Errors} \label{sec:qa-mis}
When a model predicts an answer that is not among the choices presented in the question, it can be considered an error or an out-of-scope prediction. Here are some common types of errors in this scenario:
 \ref{table:qa-predictions}
	 	\begin{table*}[h!]
		\centering
		\begin{tabular}{cp{11cm}}
			\toprule
			\textbf{Source of Mistake} & \textbf{Question, Choices, Prediction, and Correct Choice} \\
			\midrule
			\makecell{Plausible \\ Alternative} & \begin{tabular}[t]{@{}p{12cm}@{}}
				Choices: 1. \textcolor{teal}{\textbf{build trust}}, 2. hurry home, 3. ignore people, 4. believe in god, 5. jump to conclusions \\
				Question: \textbf{What do people do when networking? }\\
				Prediction: \textcolor{blue}{talk to each other} \\
				\midrule
			\textit{Comment: The prediction, while not among the choices, is a common understanding (more probable) of what people do when networking.}
			\end{tabular} \\
			\midrule
			Question fallacy & \begin{tabular}[t]{@{}p{12cm}@{}}
				Choices: 1. books, 2. \textcolor{teal}{\textbf{dard}}, 3. sky, 4. closed room, 5. television \\
				Question: \textbf{What language type is someone from Iran likely to use?} \\
				Prediction: \textcolor{blue}{Persian} \\
				\midrule
				\textit{Comment: Persian, while not a language type, is the language of Iran. Moreover, Persian belongs to the Iranian branch of the Indo-Iranian family within the Indo-European languages, while Dard refers to a language group in Pakistan. Thus, the question may be flawed.}
			\end{tabular} \\
			\midrule
			\makecell{Ambiguous \\ or misleading question} & \begin{tabular}[t]{@{}p{12cm}@{}}
				Choices: 1. summer, 2. park, 3. desktop, 4. sea, 5. \textcolor{teal}{\textbf{moon}} \\
				Question: \textbf{What blocks sunshine?} \\
				Prediction: \textcolor{blue}{cloud} \\
				\midrule
			\textit{	Comment: A cloud could be a more common answer to this question.}
			\end{tabular} \\
			\midrule
			\makecell{Synonym or  \\ different grammatical format \\ or spelling errors in question} & \begin{tabular}[t]{@{}p{12cm}@{}}
				Choices: 1.\textcolor{teal}{\textbf{ get wet}}, 2. eat vegetables, 3. falling, 4. wool sweater, 5. sharp claws \\
				Question: \textbf{The cat carefully navigated the area, they do everything they can to avoid what?} \\
				Prediction: \textcolor{blue}{getting wet} \\
				\hline
				Choices: 1. \textcolor{teal}{\textbf{skate}} 2. listen 3. opera 4. opera 5. relax \\
				Question:Punk rock music is an important part of what action sport? \\
				Prediction: \textcolor{blue}{skateboarding} \\				
				\midrule
				\textit{Comment: The prediction is essentially correct but presented in a different grammatical format or as a synonym of the correct choice, which may even fit the sentence better. In some cases, the correct choice contains misspelled words.}
			\end{tabular} \\
			\bottomrule
		\end{tabular}
		\caption{Types of sources of mistakes when the model predicts an answer not among the choices. The correct answer is shown in green, and our comments on the types of mistakes are provided in the comment section.}					
		\label{table:qa-predictions}
	\end{table*}

	\section{Details of Baselines} \label{apx:baselines}
	An overview of the related works serves as the baseline for our proposed methods for Question Answering tasks presented below.
	\begin{itemize}
		\item \textbf{QAGNN (Yasunaga et al., 2021)} \cite{base-QAGNN} employs a Graph Attention Network (GAT) (Veli?kovi? et al., 2017) to perform joint reasoning over the Commonsense Knowledge Graph (CSKG) and integrate its information into the model's processing.
		\item \textbf{GSC (Wang et al., 2021b)}  \cite{base-GSC} uses a straightforward graph neural counter as the knowledge graph (KG) encoder to incorporate knowledge from the CSKG.
		\item \textbf{GreaseLM (Zhang et al., 2022)} \cite{base-GreaseLM} integrates encoder representations from a pre-trained language model (PLM) and a KG encoder by utilizing multiple modality interaction layers, thus embedding knowledge from the CSKG into the PLM's processing.
		\item \textbf{SAFE (Jinhao Jiang and Wen, 2022)} \cite{base-SAFE} employs a multi-layer perceptron (MLP)-based KG encoder to extract features from relation paths in the retrieved multi-hop knowledge subgraph.
		\item \textbf{MVP-Tuning (Huang et. al, 2023)} \cite{mvp} utilizes comparable question-answer pairs from the training set to enhance knowledge retrieval and employs a single prompt-tuned pre-trained language model to jointly model knowledge and input text.
	\end{itemize}
\end{appendices}
\end{document}